\documentclass[letterpaper, 10 pt, conference]{ieeeconf}  
\usepackage{amsmath}
\usepackage{amsthm}
\usepackage[normalem]{ulem}
\usepackage{amssymb}
\usepackage[utf8]{inputenc}
\usepackage[T1]{fontenc}
\usepackage{graphicx}
\usepackage{enumerate}
\usepackage{xcolor}
\usepackage{comment}
\usepackage[]{biblatex}
\usepackage{multicol}
\usepackage{hyperref}
\hypersetup{
           breaklinks=true,   
           colorlinks=true,   
        }
\usepackage{tabularx}
\usepackage{makecell}
\usepackage[ruled,vlined]{algorithm2e}
\usepackage{tablefootnote}
\usepackage{multirow}
\usepackage{paralist}
\usepackage{placeins}
\usepackage{caption}
\usepackage[caption=false]{subfig}

\hypersetup{
    colorlinks=true,
    linkcolor=blue,
    filecolor=magenta,      
    urlcolor=cyan,
}

\newtheorem{proposition}{Proposition}
\theoremstyle{definition}
\newtheorem*{remark}{Remark}
\theoremstyle{definition}
\newtheorem{definition}{Definition}

\IEEEoverridecommandlockouts                              

\title{\LARGE \bf Tolerant Barrier States Embedded Trajectory Optimization for Improved Exploration  and Exploitation}

\title{\LARGE \bf Tolerant Barrier States for Improved Exploration and Exploitation  in Safety Embedded Trajectory Optimization}

\title{\LARGE \bf Tolerant Barrier States for Improved Exploration in Safety Embedded Differential Dynamic Programming}

\title{\LARGE \bf Constrained Differential Dynamic Programming Using Barrier States with Improved Exploration}

\title{\LARGE \bf Improved Exploration for Safety-Embedded Differential Dynamic Programming Using Tolerant Barrier States}

\author{Joshua E. Kuperman, Hassan Almubarak, Augustinos D. Saravanos and Evangelos A. Theodorou
\thanks{This work is supported by the National Science Foundation, CPS under Grant 1932288.} 
\thanks{Georgia Institute of Technology, Atlanta, GA, USA}
\thanks{{\tt\footnotesize \{jkup14, halmubarak, asaravanos, evangelos.theodorou\}@gatech.edu}} 
}
        
\begin{document}

\maketitle
\thispagestyle{empty}
\pagestyle{empty}

\begin{abstract}
In this paper, we introduce Tolerant Discrete Barrier States (T-DBaS), a novel safety-embedding technique for trajectory optimization with enhanced exploratory capabilities. The proposed approach generalizes the standard discrete barrier state (DBaS) method by accommodating \emph{temporary} constraint violation during the optimization process while still approximating its safety guarantees. Consequently, the proposed approach eliminates the DBaS's safe nominal trajectories assumption, while enhancing its exploration effectiveness for escaping local minima. Towards applying T-DBaS to safety-critical autonomous robotics, we combine it with Differential Dynamic Programming (DDP), leading to the proposed safe trajectory optimization method T-DBaS-DDP, which inherits the convergence and scalability properties of the solver. The effectiveness of the T-DBaS algorithm is verified on differential drive robot and quadrotor simulations. In addition, we compare against the classical DBaS-DDP as well as Augmented-Lagrangian DDP (AL-DDP) in extensive numerical comparisons that demonstrate the proposed method's competitive advantages. Finally, the applicability of the proposed approach is verified through hardware experiments on the Georgia Tech Robotarium platform. 
\end{abstract}
\section{Introduction}

As autonomous systems become more prevalent in our day-to-day lives, ensuring the safe operation of such systems is becoming of paramount importance. Proportionately, safety-critical robotics have been a center of attention for researchers recently, with applications ranging from autonomous driving and collaborative robots to aerospace applications and space exploration. Nevertheless, safety restrictions pose tremendous challenges especially when other performance objectives are also specified. A viable tool to address such challenges in a principled fashion is optimal control theory.


Trajectory optimization methods can be generally classified into two main categories. The first class contains direct approaches where the full optimal control problem is solved as a static optimization problem by a nonlinear programming solver \cite{diehl2009nonlinearmpc}. On the contrary, shooting methods parameterize only the control variables, while the states are obtained through forward integration of the dynamics \cite{von1992direct}. Differential Dynamic Programming (DDP) is a widely used second-order shooting method that has found numerous successful robotics applications in manipulation \cite{kumar2016optimal}, motion planning under uncertainty \cite{aoyama2021receding}, flight control \cite{alothman2016quadrotor}, multi-robot systems \cite{saravanos2022distributed}, etc., thanks to its convergence properties and scalability. Nevertheless, as the original DDP algorithm does not account for state constraints, encompassing safety characteristics within DDP presents significant challenges.  

Barrier based methods have shown to be a viable approach in addressing safety in control systems. Barrier certificates \cite{prajna2004safety}, barrier-Lyapunov transformation \cite{Peng2009barrier-lyap}, control barrier functions (CBFs) \cite{wieland2007constructive,ames2016CBF-forSaferyCritControl,romdlony2016stabilization} and barrier states (BaS) \cite{Almubarak2021SafetyEC,almubarak2022safeddp} all have been adopted in the literature. In safety-critical trajectory optimization especially, a recent approach that combines DDP and discrete barrier states (DBaS) \cite{almubarak2022safeddp} is shown to outperform other barrier based methods such as the quadratic penalty and CBFs. This is due to the fact that embedding barrier states into the system's dynamics eliminates the relative-degree requirement of CBF based methods while providing feedback policies of the barrier that enhance safety and robustness. 
Moreover, when combined with DDP, it enjoys standard DDP convergence analysis and allows for optimizing over the barrier state along with the system's state and control \cite{almubarak2022safeddp}. Nonetheless, despite the safety guarantees using \textit{classical} barrier functions, their use often prompts limited exploration of the state space. Specifically, only feasible solutions are allowed when searching for the optimal policy. This may limit \textit{local} iterative algorithms' ability to converge to a meaningful minimum and is prone to local minima surrounding the unsafe regions. In addition, this means that only safe initializations are allowed.

On the other hand, a significant portion of trajectory optimization methods relies on treating the constraints as soft ones by incorporating them into the cost function \cite{aoyama2020constrainedDDP, howell2019altro, jallet2022constrained}. One important advantage of such methods is that they allow for unsafe trajectory initialization, which can be crucial in complex environments where finding an initial safe trajectory is a demanding problem on its own. Additionally, they allow for intermediate solutions to be partially unsafe, which enhances their ability to arrive at non-trivial solutions. In other words, allowing for some tolerance for temporary constraint violation during the optimization process enhances the \textit{exploration} capabilities of a constrained trajectory optimization algorithm via the gradient in the unsafe set. Nevertheless, the main limitation of such \textit{tolerant} approaches lies in admitting the risk that the final solution might also be unsafe. Additionally, such methods require additional parameters that need considerable tuning.

In this work, we aim to merge the safety capabilities of DBaS-DDP and the exploration efficiency of tolerant approaches into a single methodology. Towards this direction we propose Tolerant DBaS (T-DBaS)-DDP, a generalization of DBaS-DDP that allows for temporary constraint violation while iteratively improving the solution. Our proposed method utilizes a novel barrier function shaped to approximate DBaS's safety guarantees \emph{and} have access to the gradient information within the unsafe set, avoiding local minima. We exemplify the exploratory and safe initialization advantages T-DBaS enjoys over DBaS through differential drive and quadrotor simulations as well as hardware experiments on a multi-robot platform. In addition, we demonstrate T-DBaS's advantages over state-of-the-art constrained DDP methods in performance, convergence speed, and avoiding discretization abuses. 

The remaining of this paper is organized as follows. In \autoref{PRELIMINARIES}, we introduce the constrained optimal control problem and briefly present the DBaS-DDP method. In \autoref{Section: Improved Exploration via Tolerant Barrier States}, we propose the novel tolerant barrier state approach and demonstrate the necessary derivations to formulate T-DBaS embedded DDP. Moreover, we discuss how non-vanishing T-DBaS derivatives benefit DDP in addition to guidelines for setting suitable T-DBaS parameters. Simulation results along with a comparison against related methods are provided in \autoref{Section: Applications}. Finally, the conclusions of this paper and future research directions are stated in \autoref{CONCLUSIONS}.

\section{Safety Embedded Trajectory Optimization Preliminaries} \label{PRELIMINARIES}
Consider the safety-critical finite horizon optimal control problem
\begin{equation} \label{eq: min general cost function}
    J (x_0,U) = \min_U \sum_{k=0}^{N-1} l(k, x_k, u_k) + \Phi(x_N)
\end{equation}
subject to the discrete-time, nonlinear dynamical system
\begin{equation} \label{eq: discrete control system}
    x_{k+1} = f(k,x_k,u_k)
\end{equation}
and the safety condition 
\begin{equation} \label{eq: safety condition}
h(k, x_k) > 0, \ \forall k \in [0, N]; \ x_0 \in \mathcal{S} \subset \mathbb{R}^n
\end{equation}
where $x_k \in \mathcal{X} \subset \mathbb{R}^n, u_k \in \mathcal{U} \subset  \mathbb{R}^m, U:=\{u_1,u_{2}, \dots , u_{N-1} \}, l: \mathcal{X} \times \mathcal{U} \rightarrow \mathbb{R}$ is the running cost, $\Phi: \mathcal{X} \rightarrow \mathbb{R}$ is the terminal cost, $f: \mathcal{X} \times \mathcal{U} \rightarrow \mathcal{X}$ is the dynamics describing the evolution of the system given at state $x_k$ under the control $u_k$ and $h: \mathcal{X} \rightarrow \mathbb{R}$ is a continuously differentiable function defining the safe set $\mathcal{S}:= \{x_k \in \mathcal{X} | h(x_k) >0 \}$. The goal of the safety constrained trajectory optimization is to compute the optimal control policy $U^*$ that minimizes \eqref{eq: min general cost function} such that the the safety condition \eqref{eq: safety condition} is satisfied over the whole horizon. 

\subsection{Barrier States}
By definition, guaranteeing safety is equivalent to rendering the safe set controller invariant over the horizon \cite{almubarak2022safeddp}. 
\begin{definition}  \label{invariant def}
The set $\mathcal{S} \subset \mathbb{R}^n$ is said to be controlled invariant with respect to the finite horizon optimal control problem \eqref{eq: min general cost function}-\eqref{eq: safety condition} under the feedback policy $U^*(x)$, if $\forall x_0 \in \mathcal{S}$, $x_k \in \mathcal{S} \ \forall k\in [0,N]$.
\end{definition}

To generally enforce controlled invariance of $\mathcal{S}$ using barrier functions methods, \cite{Almubarak2021SafetyEC} introduced barrier states (BaS) which are embedded into the model of the safety-critical system to be driven and stabilized with the original states of the system. In doing so, the safety objectives are expressed as performance objectives in a higher dimensional state space. In \cite{Almubarak2021SafetyEC}, the barrier state embedded model, referred to as the safety embedded model, is asymptotically stabilized, which implies safety due to boundedness of the barrier state. Next, discrete barrier states (DBaS) were proposed for discrete time trajectory optimization in \cite{almubarak2022safeddp} with differential dynamic programming (DDP). Namely, the barrier function $B: \mathcal{S} \rightarrow \mathbb{R}$ is defined over $h$. A key point in the definition of the barrier function $B\big(h(x_k)\big)$, is that $B \rightarrow \infty $ as $x_k \rightarrow \partial \mathcal{S}$. With this, a barrier function over $x$ can be defined as $\beta(x_k) := B\big(h(x_k)\big)$. In the discrete setting, the barrier state is simply constructed to be
\begin{equation} \label{eq: discrete barrier state}
    \beta_{k+1} = B \circ h \circ f(k, x_k, u_k ).
\end{equation}
For multiple constraints, multiple barrier functions can be added to form a single barrier or multiple barrier states \cite{almubarak2022safeddp}. Then, the barrier state vector $\beta \in \mathcal{B} \subset \mathbb{R}^{n_\beta}$, where $n_\beta$ is the dimensionality of the barrier state vector, is appended to the dynamical model resulting in the \textit{safety embedded system}
\begin{equation} \label{eq: safety embedded model}
    \hat{x}_{k+1} = \hat{f}(k, \hat{x}_k, u_k ), 
\end{equation}
where $\hat{x} = [x \ ; \ \beta]$ and $\hat{f} = [f \ ;\ B \circ h \circ f]$.

One of the benefits of a safety embedded model is the direct transmission of safety constraint information to the optimal controller. This prevents two separate algorithms from \textit{fighting} one another for control bandwidth, i.e. the controller attempting to maximize performance and a safety filter attempting to maximize safety. This comes at a cost of the user having to specify the weighting between task performance and safety. For the predictive control problem in this work, the following proposition \cite{almubarak2022safeddp} depicts the safety guarantees provided by the embedded barrier state method.
\begin{proposition} \label{prop:safety}
Under the control sequence $U(x)$, the safe set $\mathcal{S}$ is controlled forward invariant if and only if $\beta(x(0)) < \infty \Rightarrow \beta_k <\infty \ \forall k \in [1, T]$.
\end{proposition}
Thereby, the constrained trajectory optimization problem in \eqref{eq: min general cost function}-\eqref{eq: safety condition} is transformed to an unconstrained one in a higher dimensional space, which can be written as
\begin{equation} \label{eq: min general cost function for safety embedded system}
    J (\hat{x},U) = \min_U \sum_{k=0}^{N-1} l(k, \hat{x}_k, u_k) + \Phi(N, \hat{x}_N)
\end{equation}
subject to $\hat{x}_{k+1} = \hat{f}(k,\hat{x}_k,u_k)$.

\subsection{Trajectory Optimization Using DDP}
Differential Dynamic Programming (DDP) \cite{mayne1966second,jacobson1970differential} is a second-order trajectory optimization method that yields the optimal  locally optimal feed-forward and feedback control polices around a nominal trajectory. To enforce safety, we use the developed safety embedding methodology with DDP as done in \cite{almubarak2022safeddp}. DDP is based on dynamic programming, a.k.a. Bellman's optimality principle, which for the safety embedded optimal control problem \eqref{eq: min general cost function for safety embedded system}, states 
\begin{equation} \label{eq: Bellman eq}
V_k(\hat{x}_k)=\min_{u_k} [l(k, \hat{x}_k,u_k)+V_{k+1}(f(k,\hat{x}_k,u_k))]
\end{equation}
with boundary condition $V_N = \Phi$. Notice that the Bellman equation in \eqref{eq: Bellman eq} is for the safety embedded system and thus optimizes the barrier states along the system's states and controls \cite{almubarak2022safeddp}. Given a nominal control sequence $\bar{U}$, the algorithm iteratively solves the optimal control problem starting by expanding the perturbation of the value function to the second order around the $k^{\text{th}}$ nominal state-input pair $(\bar{\hat{x}}_k,\bar{u}_k)$ and solving \eqref{eq: Bellman eq} to find a local optimal control variation. Namely, defining 
\begin{equation*} \begin{split}
    & \hat{Q}(\delta \hat{x}_k, \delta u):= l(k, \bar{\hat{x}}_k + \delta \hat{x}_k, u_k + \delta u_k) - l(k, \hat{x}_k,u_k) + \\
    & V_{k+1}(f(k,\bar{\hat{x}}_k+\delta \hat{x}_k, u_k + \delta u+k)) - V_{k+1}(f(k,\bar{\hat{x}}_k,u_k))
\end{split} \end{equation*}
Hence, expanding both sides of \eqref{eq: Bellman eq} to the second order and matching the terms yields the Riccati-like equations
\begin{align*} \begin{split}
& V_k = V_{k+1}-\frac{1}{2} \hat{Q}_{u_k} \hat{Q}_{uu_k}^{-1} \hat{Q}_{u_k}^{^\text{T}}\ , V_{\hat{x}_k}= \hat{Q}_{\hat{x}_k} - \hat{Q}_{\hat{x}u_k} \hat{Q}_{uu_k}^{-1} \hat{Q}_{u_k} \\
& V_{\hat{x}\hat{x}_k}=\frac{1}{2} (\hat{Q}_{\hat{x}\hat{x}_k}-\hat{Q}_{\hat{x}u_k} \hat{Q}_{uu_k}^{-1} \hat{Q}_{u\hat{x}_k} )
\end{split} \end{align*}
with terminal conditions $V_N = \Phi(\hat{x}_N), \ V_{\hat{x}_N} = \Phi_{\hat{x}}(\hat{x}_N)$ and $\ V_{\hat{x}\hat{x}_N} = \Phi_{\hat{x}\hat{x}}(\hat{x}_N)$, where
\begin{align}
\begin{split} \label{eq: Q matrices}
& \hat{Q}_{\hat{x}_k}=l_{\hat{x}_k}  + {V^{^\text{T}}_{\hat{x}_{k+1}}} \hat{f}_{\hat{x}_k}, \ \hat{Q}_{u_k}= l_{u_k}  + {V^{^\text{T}}_{\hat{x}_{k+1}}} \hat{f}_{u_k} \\
&\hat{Q}_{\hat{x}\hat{x}_k}= l_{\hat{x}\hat{x}_k}  +   \hat{f}_{\hat{x}_k}^{^\text{T}} V_{\hat{x}\hat{x}_{k+1}} \hat{f}_{\hat{x}_k}  \\ 
&\hat{Q}_{uu_k}=  l_{uu_k}  +  \hat{f}_{u_k}^{^\text{T}} V_{\hat{x}\hat{x}_{k+1}} \hat{f}_{u_k}   \\
&\hat{Q}_{\hat{x}u_k}=  l_{\hat{x}u_k}  +  \hat{f}_{\hat{x}_k}^{^\text{T}}  V_{\hat{x}\hat{x}_{k+1}} \hat{f}_{u_k} 
\end{split}
\end{align}
Then, the optimal control variation that minimizes the expanded Bellman's equation is
\begin{equation} \label{eq: optimal var u}
\delta u^*_k = - \hat{Q}_{uu_k}^{-1} \big( \hat{Q}_{u_k}^{\rm{T}} +  \hat{Q}_{u \hat{x}_k} \delta \hat{x}_k\big) = {\rm{\mathbf{k}}}_k + {\rm{\mathbf{K}}}_k \delta \hat{x}
\end{equation}
Once the backward pass is completed, the forward pass propagates the safety embedded system \eqref{eq: safety embedded model} using the optimal control sequence computed using \eqref{eq: optimal var u}. Then, the new trajectory is linearized around and the process is repeated until convergence. In addition, we use the improved DDP in \cite{tassa2012synthesis}, with regularization and line-search.


A very important assumption in barrier based methods is the requirement for the nominal trajectory to always be safe, thus also necessitating a safe initialization. Those assumptions guarantee safety using the fact that DDP can guarantee improvement of optimality of the trajectory, i.e. cost reduction \cite{almubarak2022safeddp}. Following this, we relax the requirement for a safe nominal trajectory by allowing for temporary constraint violations which also encourages exploration and avoiding local minima.



\section{Improved Exploration via Tolerant Barrier States} \label{Section: Improved Exploration via Tolerant Barrier States}
\subsection{Tolerant Barrier States}

T-DBaS are motivated to introduce more flexibility during the optimization process by allowing for temporary constraint violations during solver iteration. Such a behavior is particularly useful in several scenarios and can greatly improve the optimization algorithm capabilities.
An example is shown in \autoref{fig: corridor example}, where a vehicle is required to reach a target behind a wide wall. Due to using the classical barrier, standard DBaS-DDP exhibits limited exploration capabilities and cannot find a safe trajectory that reaches to the target. On the contrary, allowing for temporary violations enhances exploration efficiency. Several examples in which such scenarios happen are discussed in \autoref{Section: Applications}.

Nonetheless, since theoretical safety guarantees would not hold anymore,
the barrier should be designed such that its value sharply and continuously increases inside the unsafe regions. 
Indeed, note that the barrier's derivatives are being utilized in the barrier state's dynamics and cost. This means that the barrier's gradient should point outside 
the unsafe region with an increasing intensity instead of diminishing quickly from infinity as is the case for classical barriers (see \autoref{fig: tolerant barrier plots for illustrations}). In addition, the Hessian should always remain positive definite to ensure proper minimization. 

To achieve these objectives, we propose a new tolerant barrier, $\tilde{B}$, where the desired characteristics are acquired by summing the sigmoid and the softplus function. Specifically, for the safety function $h$, the corresponding sigmoid function is defined as
\begin{equation*}
    \sigma(h) = \frac{1}{1 + e^{c_1h}}
\end{equation*}
where increasing $c_1$ causes $\sigma(h)$ to closely approximate the unit step function with $\frac{\partial \sigma(h)}{\partial h} = c_1 \sigma (h) \big(\sigma(h) -1\big)$.
Similarly, the softplus function is defined as
\begin{equation}
    \sigma^{+}(h) = \frac{1}{c_2} \log(1 + e^{-c_2h}) 
\end{equation}
where increasing $c_2$ causes $\sigma^{+}(h)$ to closely approximate the ReLU function with $    \frac{\partial \sigma^{+}(h)}{\partial h} =    \frac{-1}{1 + e^{c_2h}}$.
Hence, the constructed barrier function over the safety function takes the form
\begin{equation} \label{eq: tolerant barrier function}
    \tilde{B} = p \sigma(h) + m \sigma^{+}(h)
\end{equation}
with $\tilde{B}_x = \Big(p \frac{\partial \sigma(h)}{\partial h} + m \frac{\partial \sigma^{+}(h)}{\partial h} \Big) \frac{\partial h}{\partial x}$, 
where $p$ and $m$ are both scaling parameters. $p$ is the height of the sigmoid function and can be intuitively thought of as the penalty incurred by approaching the boundary into the unsafe set, and $m$ affects the gradient of $\tilde{B}$ within the unsafe region and intuitively serves as the addition penalty for traveling further into the unsafe set. 

Note that $\tilde{B}$ is differentiable and does not \textit{blow up} as we approach the boundary of the unsafe set but instead changes dramatically (high rate of change) and has non-vanishing gradient and Hessian, which will be of a great importance when used to form barrier states. Because $\tilde{B}$ value does not go to infinity at $h=0$, \autoref{prop:safety} does not apply; however, $\tilde{B}$ can be designed such that it ensures safety similary to a classical barrier. 

\begin{remark}
$\tilde{B}$ can be designed to mimic the classical barrier function $B$ near the boundaries practically enough, i.e. $\tilde{B}$ penalizes values of $h$ close to zero as aggressively or more aggressively than $B$. In other words, we can design $\tilde{B}$ such that for a bounded solution of $\beta(k)$, $\tilde{B}(h)\approx B(h)$ in the domain $\{h|h>h(t),t=\underset{k} {\arg\!\max} \beta(k)\}$ which then by \autoref{prop:safety}, guarantees safety. Additionally, inside the unsafe set $\tilde{B}$ is constantly decreasing as $h \rightarrow \partial \mathcal{S}$,  which pushes constraint violating state trajectories into the safe set, and $\lim_{h \to -\infty} \tilde{B}_x=-m$, a tunable parameter. This can be seen in \autoref{fig: tolerant barrier plots for illustrations} which shows a possible behavior of the tolerant barrier compared to some known barrier functions such as the inverse barrier and the log barrier, which both have undesirable properties where $h<0$.
\end{remark}





We are now in position to develop the tolerant discrete barrier state and embed it into the trajectory optimization problem. Define $\tilde{\beta}_k(x):= \tilde{B}\circ h(x_k)$. Then, the tolerant discrete barrier state (T-DBaS) is given by
\begin{equation} \label{eq: tdbas equation}
    \tilde{\beta}_{k+1}(x):= \tilde{B}\circ h \circ f(k,x_k,u_k)
\end{equation}
It is worth mentioning that one can design a single barrier state or multiple barrier states for multiple constraints \cite{almubarak2022safeddp}. Defining $\hat{x} = [x \ ; \ \tilde{\beta}]$ and augmenting the system \eqref{eq: discrete control system} with the T-DBaS dynamics \eqref{eq: tdbas equation} $\hat{f} = [f \ ;\tilde{B}\circ h \circ f(k,x_k,u_k)]$ transforms the problem to 
\begin{equation} \label{eq: min cost function for tdbas augmented system}
    J (\hat{x},U) = \min_U \sum_{k=0}^{N-1} l(k, \hat{x}_k, u_k) + \Phi(N, \hat{x}_N)
\end{equation}
subject to $\hat{x}_{k+1} = \hat{f}(k,\hat{x}_k,u_k)$.

\subsection{Gradient and Hessian of T-DBaS within DDP}
\begin{figure}[t]
\vspace{-5mm}
    \centering
    
    \includegraphics[trim=20 20 40 0, width=0.6\linewidth]{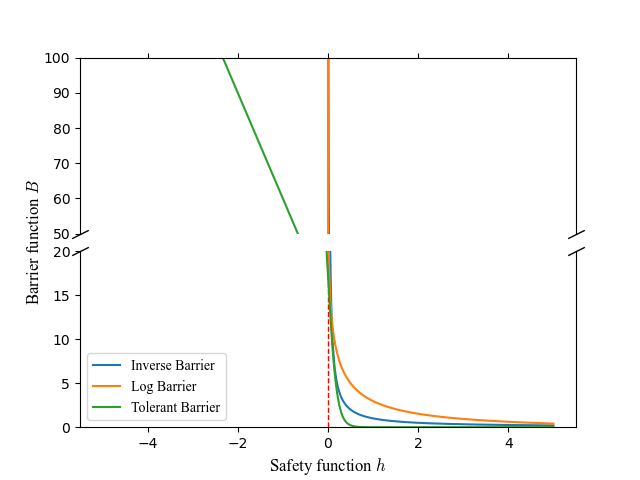} \\\hspace{-2cm}
    \includegraphics[trim=2cm 2.5cm 3cm 0cm, clip, width=0.5\linewidth]{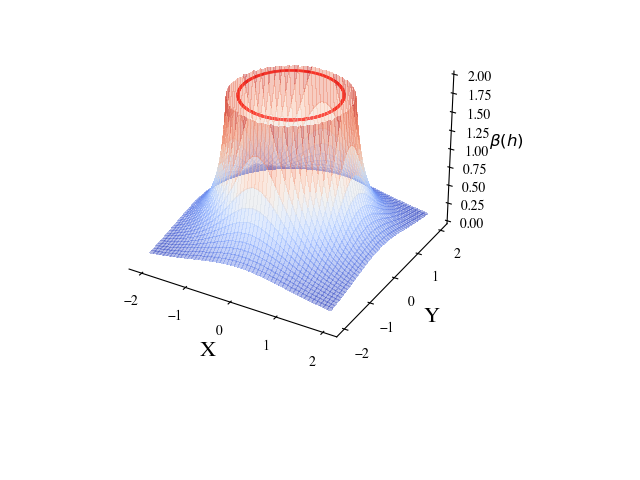} \hspace{-5mm}
    \includegraphics[trim=2cm 2.5cm 3cm 0cm, clip,width=0.5\linewidth]{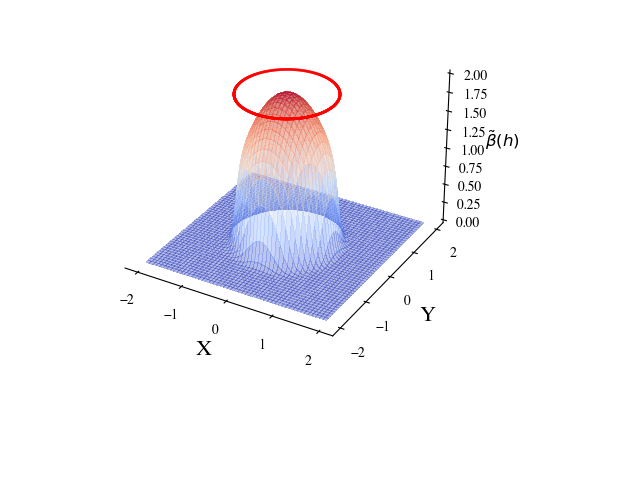} \hspace{-20mm}
    \caption{The top figure shows an example of the tolerant barrier with the inverse and log barriers. The bottom figures show the barrier functions in 3D to better illustrate their gradients. The inverse barrier (left) and the tolerant barrier (right) treat the unsafe region (circle of $r=1$) differently with the z-axis denoting the value of the barrier. Note that the discontinuity on the classical barrier function causes numerical instability when presented with unsafe initial conditions, whereas the tolerant barrier smoothly penalizes unsafe starts while also approximating the classical barrier in the safe region.  
    }
\vspace{-5mm}
\label{fig: tolerant barrier plots for illustrations}
\end{figure}

Retaining barrier information in the unsafe region helps DDP find non-trial solutions. 
To help illustrate it's benefits, \autoref{fig: tolerant barrier plots for illustrations} shows a possible design of the tolerant barrier along with the inverse barrier and the log barrier in the top sub-figure. In the bottom sub-figures, an ellipsoid representing an unsafe region is shown with the inverse barrier gradient (left) and the tolerant barrier gradient (right) represented by a field of arrows that indicate the direction and the magnitude of the gradient in the state space. The color map is added to indicate the value of the barrier. It can be seen that the proposed barrier has an intense field of arrows inside the unsafe region and at the boundaries pointing outside, unlike the classical barrier whose gradient points away from the unsafe region only at the boundaries with the ones inside weakly pointing inside the unsafe region.

One of the advantages of DBaS-DDP is its self-regularization of the quadratic approximation matrices in \eqref{eq: Q matrices}. More specifically, for the cost function in \eqref{eq: min general cost function for safety embedded system}, let the DBaS penalization matrix be $p_w \succ 0$ and let us consider the $\hat{Q}_{uu}$ matrix given previous time step information, whose positive-definiteness is needed to converge to a minimum and compute the optimal solution
\begin{align} \begin{split}
    \hat{Q}_{uu} & =  l_{uu}  +  \hat{f}_{u}^{^\text{T}} V_{\hat{x}\hat{x}} \hat{f}_{u} \\
    & = l_{uu} + f_u^T V_{xx} f_u + f_u^T h_x^T B' p_w B' h_x f_u \\
    & = Q_{uu} + f_u^T H_{\text{DBaS}} f_u
\end{split}\end{align}
where $Q_{uu}$ is for the unconstrained case and $H_{\text{DBaS}}:= h_x^T B'^T p_w B' h_x$ is the Hessian of the barrier states. Notice that this means $f_u^T H_{\text{DBaS}} f_u \succeq 0$ helps ensure proper minimization, unlike other penalty methods in which indefinite terms appear. For T-DBaS, this is especially advantageous in that the self-regularization of $Q_{uu}$ is retained even in the unsafe region, since it always has a positive definite Hessian with a steady state value of $m$, unlike classical barriers. Note that this phenomenon also appears in $\hat{Q}_{\hat{x}\hat{x}_k}$ and $\hat{Q}_{\hat{x} u}$.

\subsection{Tuning Discussion}
While T-DBaS enjoy many advantages over previous barrier methods, this comes with a trade-off of introducing new tuning parameters.
 Despite this, we have found that, with some understanding, the optimal for each parameter can be intuitively set 
based on the application requirements and the properties of the sets $\mathcal{S}$ and $\mathcal{S}^C$. 

\begin{itemize}
\item $Q_{tdbas}$: This parameter is the running cost matrix element that penalizes the barrier state. Without it, obstacles and unsafe regions are ignored. 
In general, it is best to set $Q_{tdbas}$ prior to tuning the rest of the parameters and alter it minimally, unless you find a desirable barrier function shape and want to holistically penalize the safety constraints more or less.

\item $c_1$: High values of $c_1$ make the sigmoid term of the barrier approximate the unit step function while lower values make the sigmoid smoother. This property is important to adjust how much penalty is incurred in the safe region, since a smoother sigmoid has larger values on its lower half.  If one would seek less conservative behavior, then $c_1$ can be increased before $p$. 

\item $c_2$: High values of $c_2$ make the softplus term of the barrier approximate the ReLU function while lower values make the softplus smoother. This alters the penalty in the safe region, but also how quickly the gradient of the softplus approaches $1.0$ in the unsafe region.  If $p=0$, then a higher $c_2$ is recommended to begin penalizing any small negative values of $h$. 

\item $p$: The height of the sigmoid function defines the penalty for crossing into the unsafe region. The higher $p$ is set, the closer T-DBaS will approximate classical DBaS. For problems where the obstacle or unsafe set is small, large values of p also require large $c_1$ values as to avoid overly-conservative behavior. If $c_1$ is low, then $p$ also affects the barrier values in the safe region. 

\item $m$: This parameter defines the slope of the barrier in the unsafe region.
 , or the penalty for going further into the unsafe region. 
The level of exploration that the solver enjoys is highly dependent on this parameter. With lower $m$, there is less penalty incurred for exploring the unsafe set. With higher $m$, the solver is incentivized to exit the unsafe set as quickly as possible. 
 If $c_2$ is low, then $m$ also affects the barrier values in the safe region. For problems where the obstacle or unsafe set is small, often high values of $m$ are enough to avoid collision and $p$ can be set to $0$. 
\end{itemize}

\begin{figure} [htb]
    \centering
 \vspace{-5.5mm}
     \hspace*{-0.5cm}\subfloat[Iteration 1]{\includegraphics[trim=50 0 50 40, clip, width=.45\linewidth]{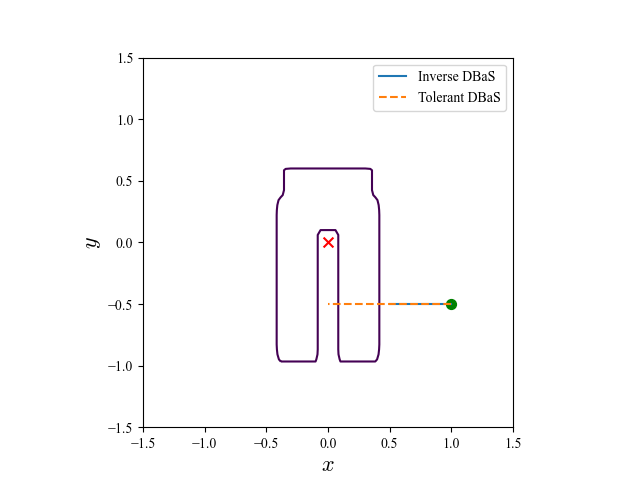}}
     \hspace*{-2mm}\subfloat[Iteration 2]{\includegraphics[trim=50 0 50 40, clip, width=.45\linewidth]{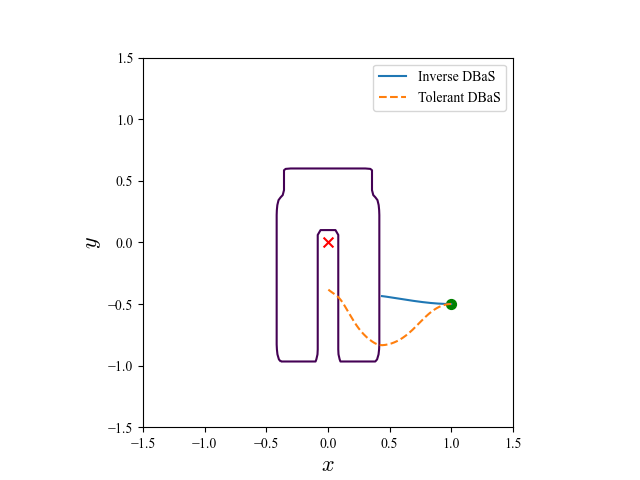}} 
     \\
     \hspace*{-0.5cm}\subfloat[Iteration 6]{\includegraphics[trim=50 0 50 40, clip, width=.45\linewidth]{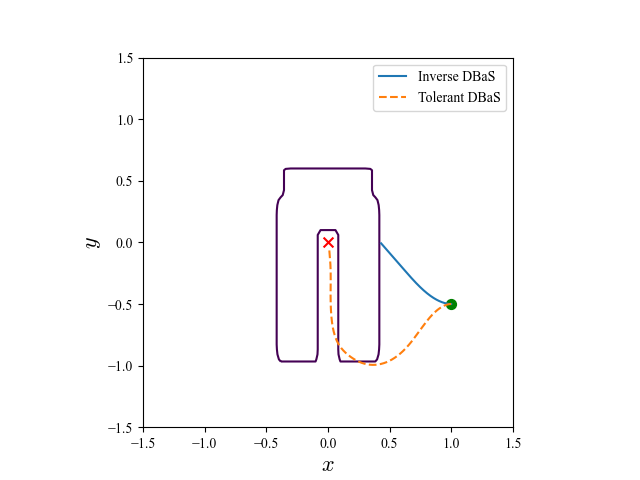}}
     \hspace*{-2mm}\subfloat[Iteration 9]{\includegraphics[trim=50 0 50 40, clip, width=.45\linewidth]{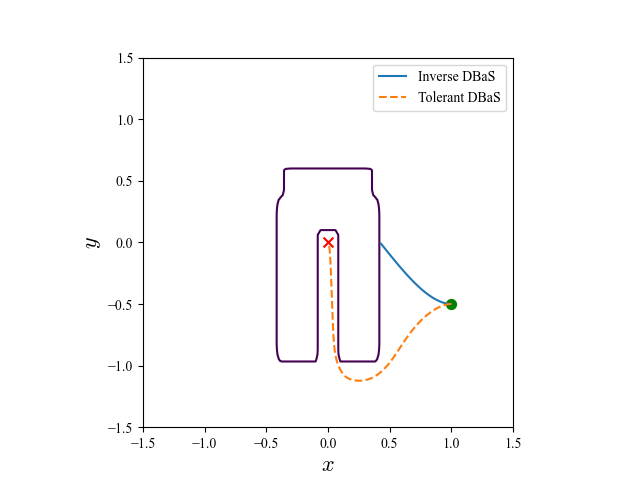}}
    \caption{Intermediate solutions for T-DBaS-DDP and DBaS-DDP. The former method avoids the local minima by taking advantage of the constraint gradient in the unsafe set, while the latter is unable to provide a trajectory that reaches to the goal.}
    \label{fig: corridor example}
     \vspace{-5mm}
\end{figure}

\section{Applications} \label{Section: Applications}
In this section, we implement T-DBaS-DDP to solve a variety of highly non-convex trajectory optimization problems that require a high degree of exploration or have unsafe initializations. We compare against DBaS-DDP \cite{almubarak2022safeddp} and AL-DDP \cite{aoyama2020constrainedDDP}. We first implement TDBaS-DDP on a differential drive system tasked with discovering the path to the goal and avoid local minima that the classical DBaS is susceptible to. The second example is a quadrotor that starts inside a hazardous region which it should leave and then track a dangerous trajectory. The third example is a multi-agent control application in which two teams of mobile robots must form (close the distance between them) and then stay formed while avoiding obstacles on their way to their targets. This is then implemented on hardware in the Georgia Tech Robotarium \cite{wilson2020robotarium}. Finally, we compare T-DBaS-DDP with DBaS-DDP and AL-DDP in navigation tasks for a differential drive robot in randomly generated fields of obstacles and for a quadrotor in randomly generated fields of moving obstacles, shown in \autoref{fig:obstacle examples}. In this section, a quadratic cost function of the form $J=\sum^{N-1}_{i=k} \hat{x}_k^{\rm{T}} Q \hat{x}_k + u_k^{\rm{T}} R u_k + \hat{x}_N^{\rm{T}} S \hat{x}_N $ is used. The differential drive robot dynamics used here is similar to that in \cite{almubarak2022safeddp} and the $12$ dimensional quadrotor model is derived in \cite{Sabatino2015QuadrotorCM}.

\subsection{Differential Drive Robot}

To show that the proposed T-DBaS enables DDP to avoid the local minima that DBaS is susceptible to, we set up a navigation problem in which the target is behind a large wall and the robot can narrowly move to get to the target with poor initialization (see \autoref{fig: corridor example}). To set up this environment, the safety functions are $h_1(x)=|3(p_x-0.5)+0.5p_y|+|3(p_x-0.5)-0.5p_y|-1$, $h_2(x)=|p_x+2(p_y-0.75)|+|p_x-2(p_y-0.75)|-1>0$, and $h_3(x)=|3(p_x+0.5)+0.5p_y|+|3(p_x+0.5)-0.5p_y|-1$ each representing rectangles that together form a horseshoe-like shape. Our barrier states for the inverse DBaS and T-DBaS methods, respectively, are then $\beta(x)=\sum_{i=1}^3\frac{1}{h_i(x)} $ and $\tilde{\beta}(x)=\sum_{i=1}^3\tilde{B}(h_i(x))$ with parameters $p=500$, $m=500$, $c1=30$, $c2=50$. The cost function's parameters are $Q=\text{diag}(0,0,0,10^{-5})$, $R = \text{diag}(0.001, 0.001)$, and $S=\text{diag}(1000,1000,0,0.05)$. The robot's initial position is $(1,-0.5)$ with the goal $(0,0)$ being surrounded by obstacles. The DDP algorithms are initialized with zero controls, i.e. not moving.


As shown in \autoref{fig: corridor example}, only the agent augmented with a tolerant barrier state reaches the goal safely while classical DBaS struggles to find a solution given the poor initialization. This is mainly due to the local nature of the solver. DBaS-DDP could not find a local improved solution that helps it avoid the obstacles as it does not accept solutions outside the safe region. Therefore, DBaS-DDP does not have enough gradient information that points towards the optimal minimum and gets stuck in a local minimum. On the other hand, T-DBaS-DDP accepts unsafe solutions and uses the high cost and gradient information to improve the solution and eventually find a safe solution. It is worth mentioning that using the tolerant barrier function's cost only in this example is insufficient to avoid the obstacles, similar to the findings in \cite{almubarak2022safeddp}, in which it is shown that the penalty method generally struggles to find safe solutions.


\subsection{Quadrotor}
In this example, a quadrotor starts in an \textit{unsafe} region, portrayed by a spherical obstacle, and is required to exit the unsafe region and track an unsafe trajectory forming a figure eight as in \cite{almubarak2022safeddp}. 
To perform the task, we construct a single T-DBaS to represent the obstacles by summing the corresponding tolerant barrier functions. 
Note that the quadrotor has to avoid the same unsafe region it started in to show the efficacy of the proposed approach in approximating the safe-set invariance of the DBaS approach despite allowing for unsafe initializations. 
The cost parameters are $Q=\text{diag}(0,\cdots, 200, 200, 200)$, $R=0.5 \cdot 10^{-2} I_{4 \times 4}$ with the running cost now defined based on the error of being away from the desired trajectory and the T-DBaS parameters are $p=10$, $m=5$, $c_1=5$, and $c_2=5$. The obstacle constraint is $h=||p-o||^2-r^2>0$, where $p$ is the quadrotor's position, $o$ is the coordinates of the obstacle's center, and $r$ is its radius. 

\begin{figure} []
 \vspace{-5mm}
   \centering
    \hspace{-4mm} \includegraphics[trim=25 25 25 25, width=.5\linewidth]{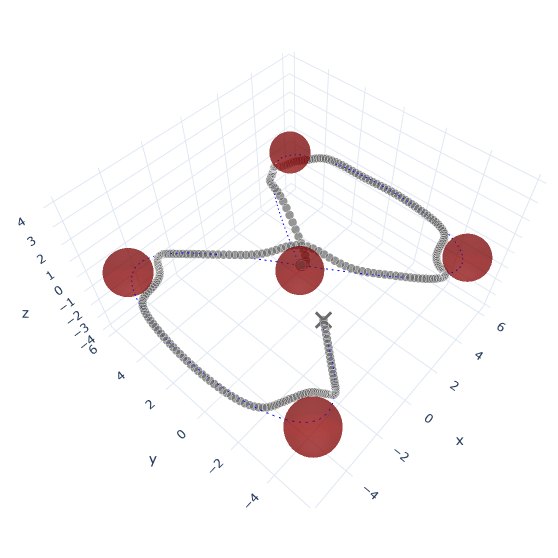}
    \includegraphics[trim=25 25 25 25, width=.5\linewidth]{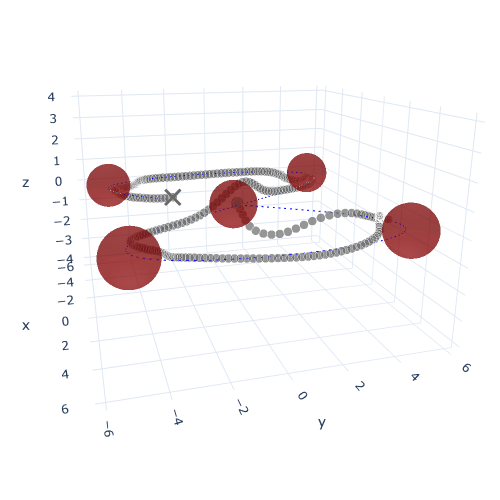} \\
    \includegraphics[trim=25 25 25 25, width=.6\linewidth]{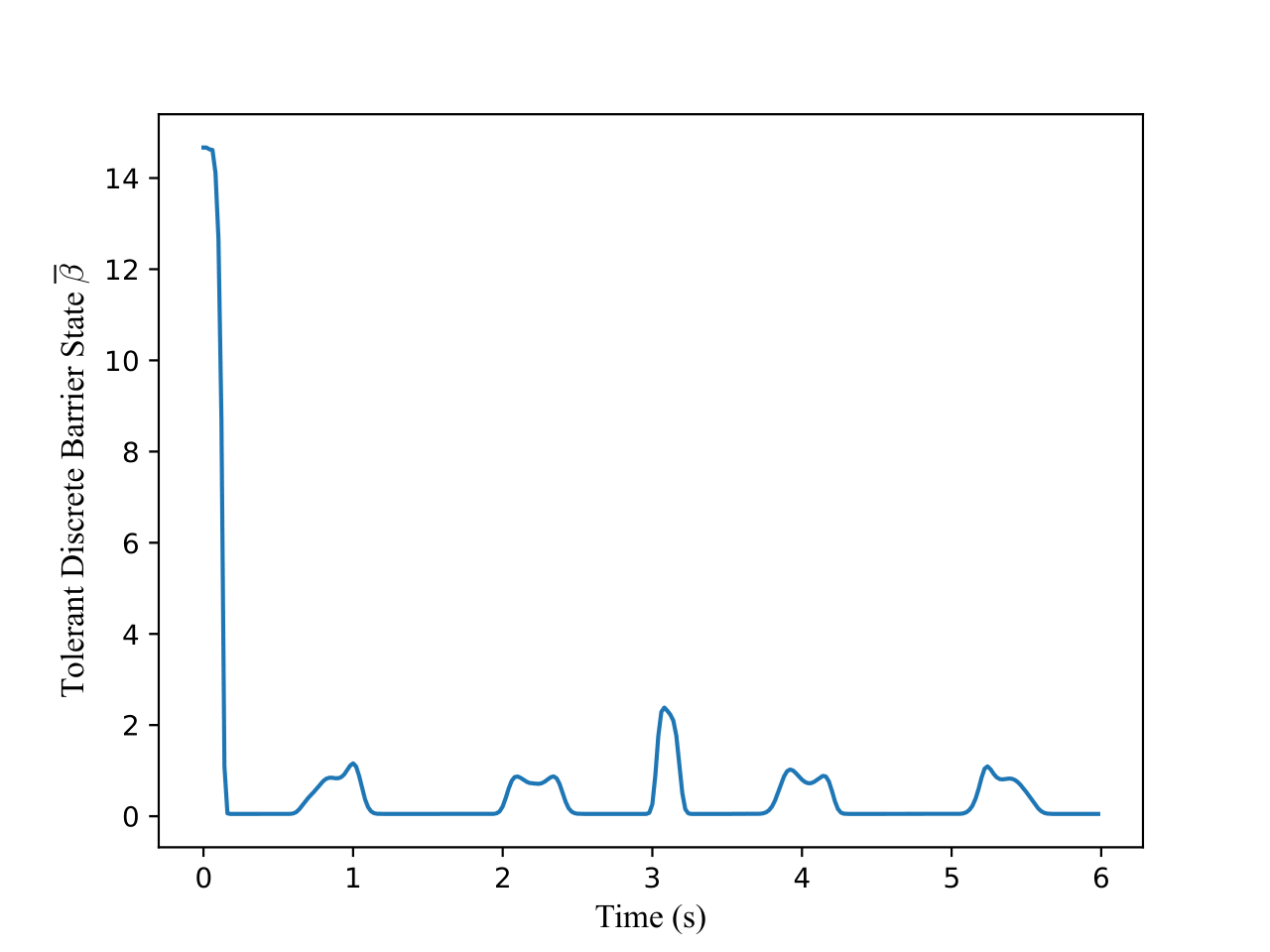}
    \caption{Two view angles of the quadrotor (filled gray circles) starting inside an unsafe region (dark red) at $(0,0,0)$ and exiting it to track an unsafe trajectory (dotted blue) and avoid obstacles on the way.}
\label{fig: t-dbas progression for quadrotor example}
\vspace{-5mm}
\end{figure}

\autoref{fig: t-dbas progression for quadrotor example} shows the quadrotor successfully exiting the unsafe region and then tracking the reference trajectory while avoiding the obstacles including the one it started in. The bottom figure shows the  progression of the T-DBaS starting with a relatively large value and decreasing as the quadrotor leaves the obstacle and increasing in value as the quadrotor travels close to the boundaries of the obstacles to track the desired trajectory.

\subsection{Teams of Mobile Robots Hardware Experiment}
In this example, two teams of two Georgia Tech Robotarium \cite{wilson2020robotarium} mobile robots have the following objectives:
\begin{enumerate}
    \item Establishing connectivity within their team despite starting far away from each other, i.e. moving within a specified distance from the other member of their team,
    \item Reaching their respective goal states,
    \item Avoiding collisions with each other and an obstacle,
    \item Staying within the Robotarium boundaries.
\end{enumerate}

The goal of this experiment is to demonstrate the performance of the resulting feedback policy of T-DBaS-DDP on a group of real-world, noisy systems. The time given to accomplish these tasks is approximately 40 seconds discretized by $dt=0.033$s with $N=1213$. 
For each objective, we construct a distinct barrier state, resulting in four barrier states for agents' connectivity, agents' collision, obstacle avoidance and staying within the Robotarium boundaries. This results in a state space of $n=12$ and a control space of $m=8$. For each barrier, the barrier values of each constraint are summed. The connectivity constraints are defined by $h^{\vartheta}_{ij}= -||x_i-x_j||^2+\vartheta^2$, where $x_i$ and $x_j$ are the positions of agents $i$ and $j$ and $\vartheta =0.3$. 
The collision constraints of all the agents are similarly calculated with $h^{\delta}_{ij}=||x_i-x_j||^2-\delta^2$ and $\delta=0.15$. The obstacles are defined as squares with $h=|p_{x}+p_{y}|+|p_{x}-p_{y}|-0.9-r_{r}>0$, where $p_x,p_y$ are the relative distances of the agents from the obstacle centers at $(0,0),(0,1.15),(0,-1.15)$. The obstacle safety function has an additional $-r_{r}$ term to account for the radius of the agent, $0.075$. The boundary safety function is defined as the minimum signed distance from the boundaries at $x=(-1.6,1.6),y=(-1,1)$, also with a $-r_{r}$ term. 

The cost parameters were chosen as $Q=\text{diag}(0.0001, \cdots, 0.1, 0.1, 0.1, 0.1)$, $R= 5 I$, and $S_a=\text{diag}(1000,1000)$ for each agent's position and $S_{\tilde{\beta}}=\text{diag}(0.05, 0.05, 0.05, 0.05)$ for the barrier states. The tolerant parameters for each constraint were chosen to be collide: $p,m,c_1,c_2=(0,100,1,175)$, connect: $p,m,c_1,c_2=(0,5,1,45)$ with $c_1=1$ to avoid dividing by zero, obstacles and boundaries: $p,m,c_1,c_2=(3,1,50,100)$. We found that in order to create a stable trade-off between connectivity and collision avoidance, a plotting of the distance, $d_{ij}$, between two agents against $\tilde{B}_{collide}+\tilde{B}_{connect}$ was necessary to design for the minimum to fall within the safe region $d_{ij}\in(0.15,0.3)$ with smooth penalizations in either direction. Specifically, with $d_{connect}=0.4$, the optimal tolerant parameters are then $c_{2,collide}=100$ and $c_{2,connect}=35$. Following this pattern for $d_{connect}=0.5$, $c_{2,connect}=25$ and $c_{2,collide}=100$ were the optimal parameters. This visualization also showed that non-zero values of $p$ pushed the minimum away from the safe region and resulted in over-correcting by the local DDP/iLQR algorithms, as the initial $\tilde{\beta}_{connect}$ is already quite high and additional penalization by $p$ was not necessary. When the safe region is larger, as in previous examples, or $d_{ij}$ is closer to it's safe range at initialization, $p$ becomes a useful asset in pushing the trajectory into the safe region and keeping it there. 

\begin{figure}[ht]
 \centering
 \subfloat[t = 0 sec]{\includegraphics[trim=0 0 0 0, clip,width=0.35\textwidth]{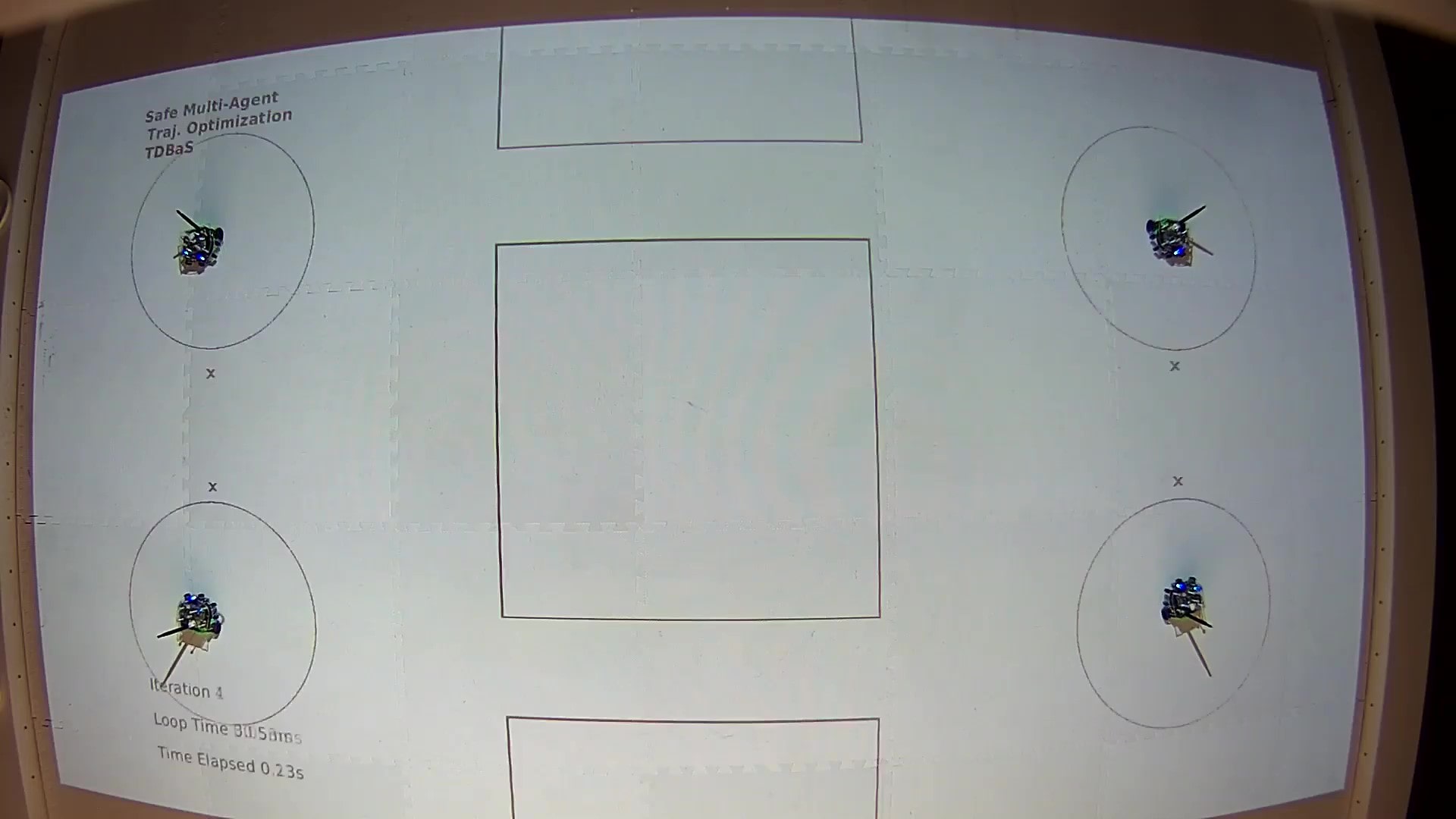}}
 \vspace{-3mm}
 \\
 \subfloat[t = 5 sec]{\includegraphics[trim=0 0 0 0, clip,width=0.35\textwidth]{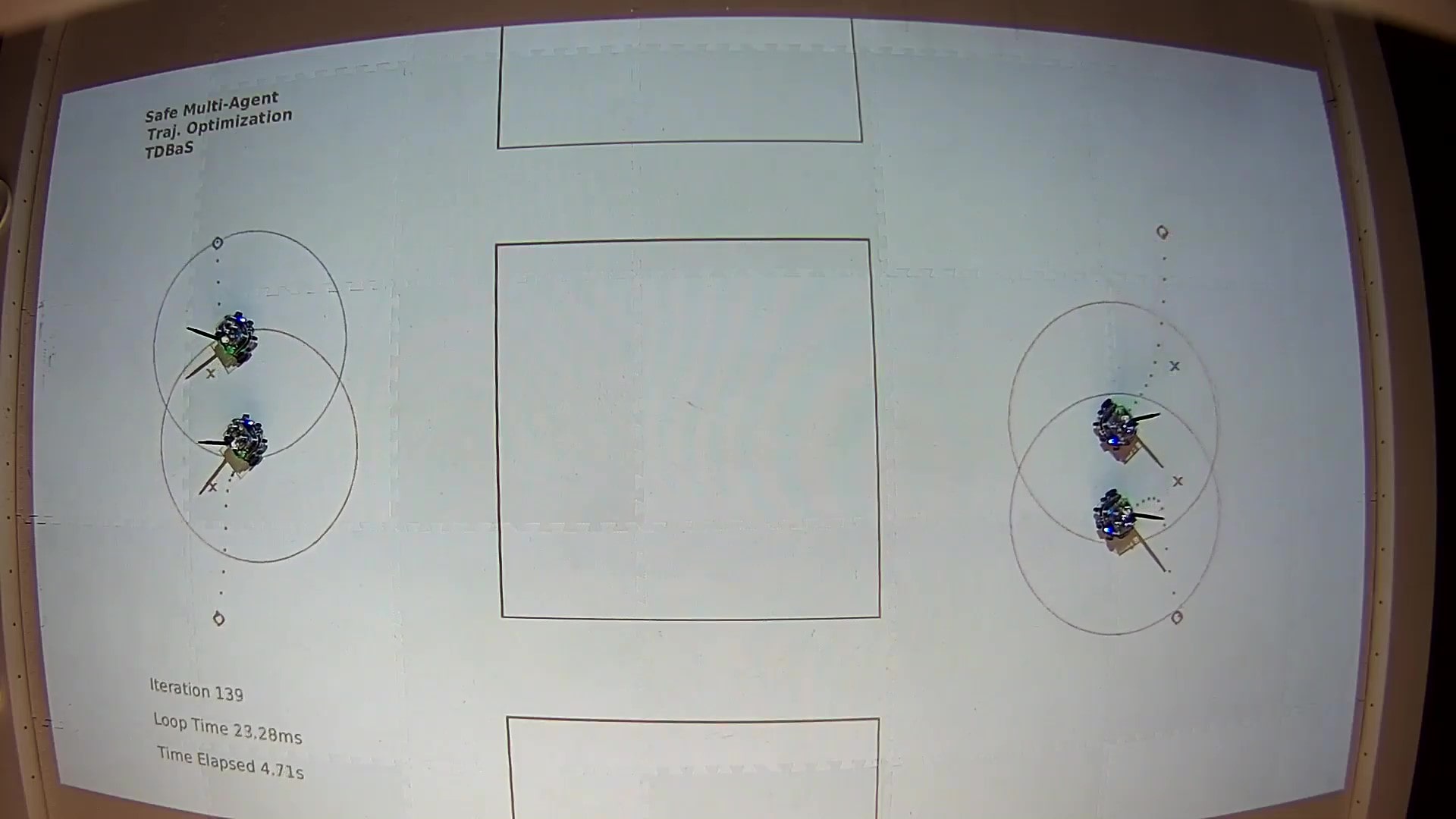}} 
 \vspace{-3mm}
 \\
 \subfloat[t = 25 sec]{\includegraphics[trim=0 0 0 0, clip,width=0.35\textwidth]{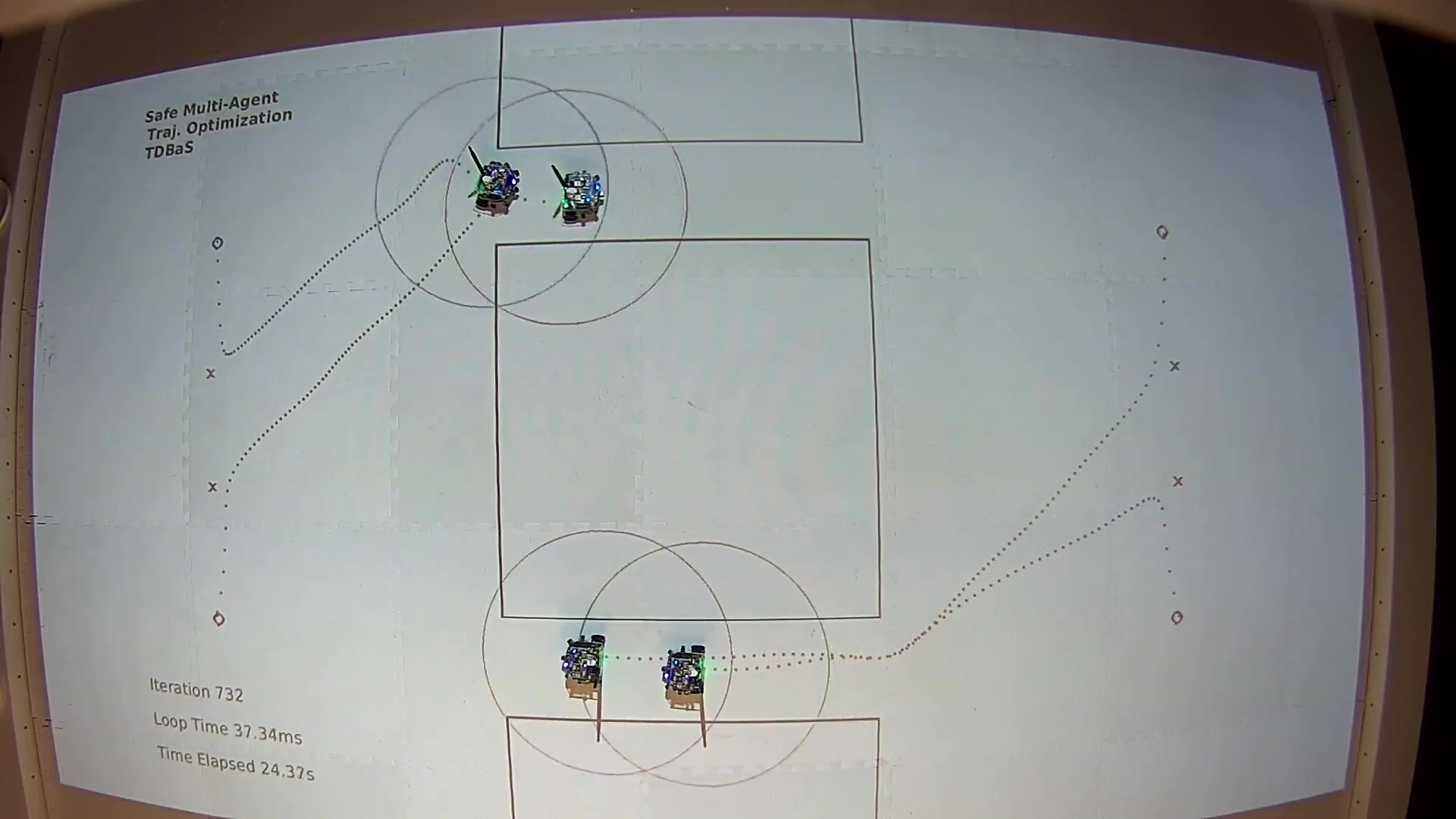}} 
 \vspace{-3mm}
 \\
 \subfloat[t = 40 sec]{\includegraphics[trim=0 0 0 0, clip,width=0.35\textwidth]{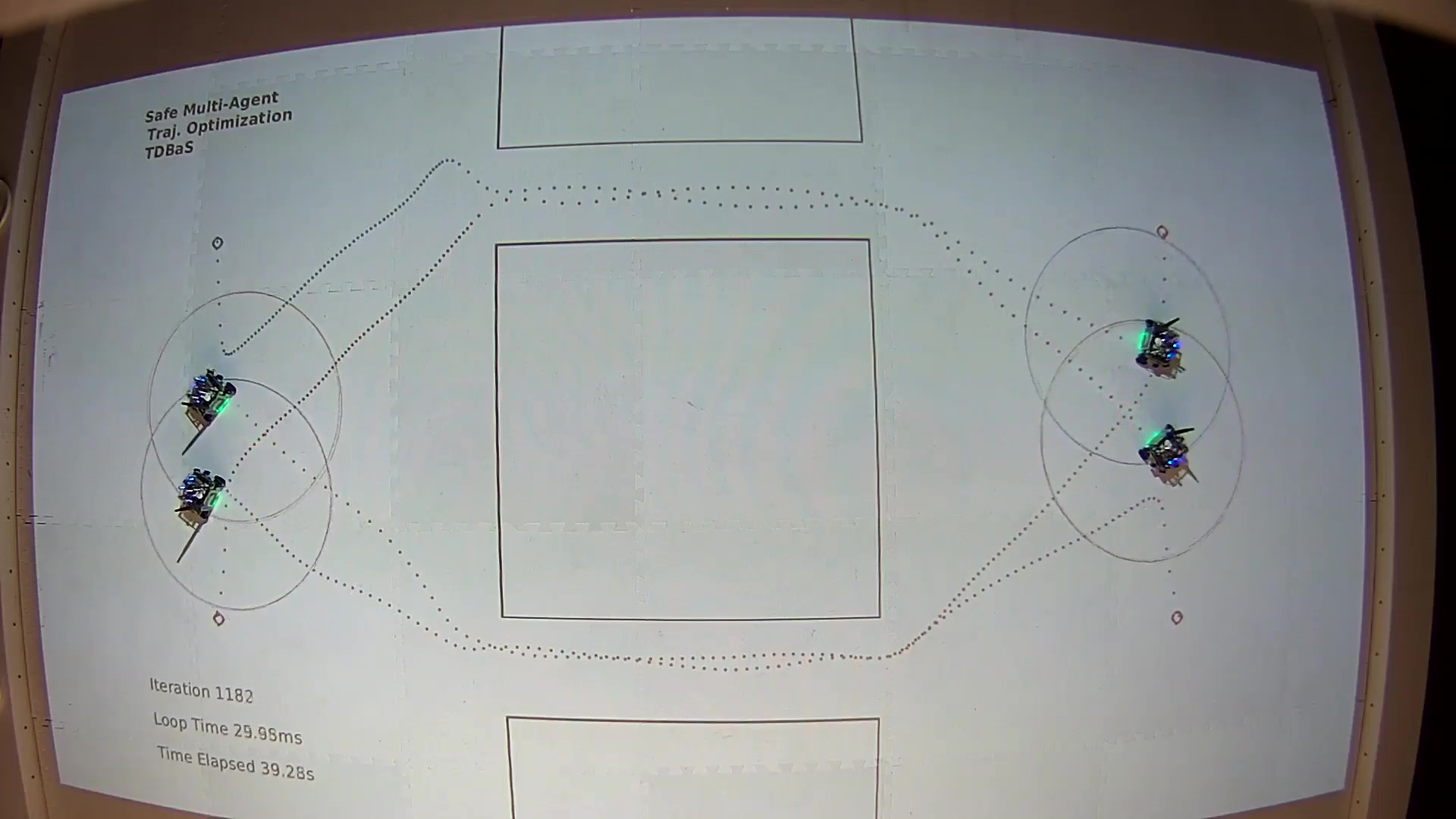}}
 \caption{Experiment of two teams of two agents per team switching positions and maintaining connectivity with their teams while avoiding obstacles using Georgia Tech Robotarium. The circles around the agents indicate the connectivity area of the agent and the dotted lines indicate the traveled trajectories. A video of the experiment can be found in https://youtu.be/9ZRBHZfjKPY. }
 \label{fig: robotarium of 4 agents}
 \vspace{-5mm}
 \end{figure}

This problem is clearly a difficult one due to the multiple constraints and the shape of the obstacles. \autoref{fig: robotarium of 4 agents} shows the successful implementation. The robots start moving fast towards each other to establish connectivity and then move together to maintain connectivity towards the target while keeping a distance and avoiding the obstacles and the other agents. The centralized approach allows T-DBaS-DDP to find a solution in which each team moves through a separate corridor to avoid collision and reach the target.

\subsection{Comparisons with Augmented Lagrangian}

\begin{figure}[!htb]
\minipage{0.21\textwidth}
  \includegraphics[trim={0 -1.5cm 0 0},clip,width=\linewidth]{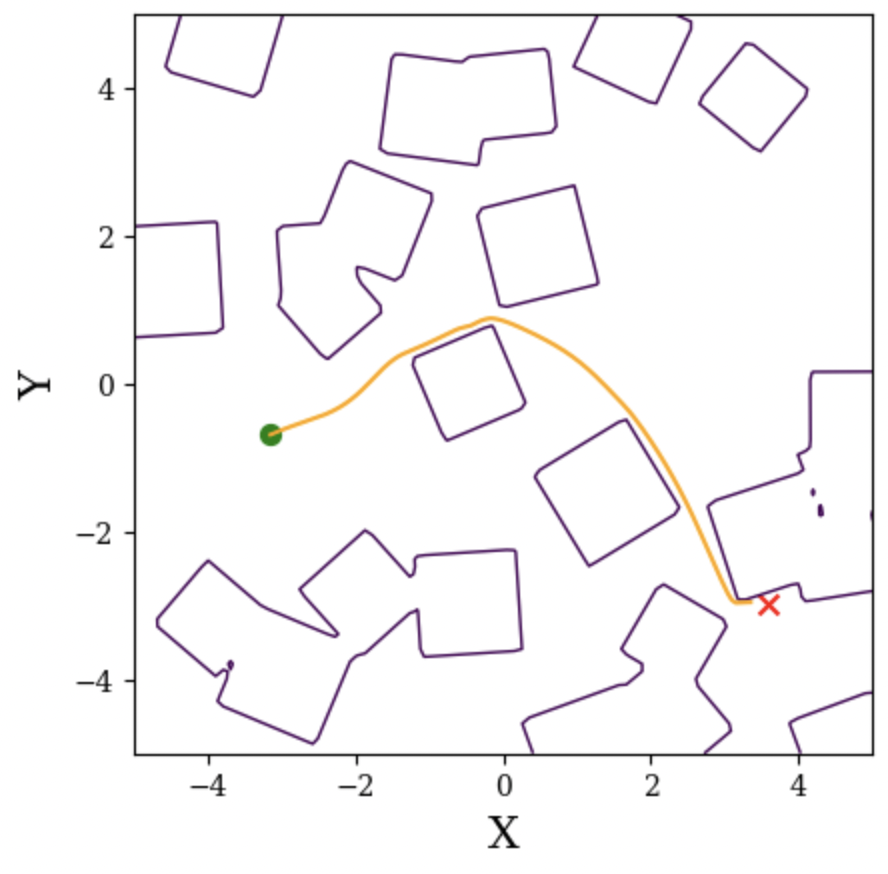}
  \caption*{a) Diff. Drive}
\endminipage\hfill
\minipage{0.27\textwidth}
  \includegraphics[trim={0 0 0 0},clip,width=\linewidth]{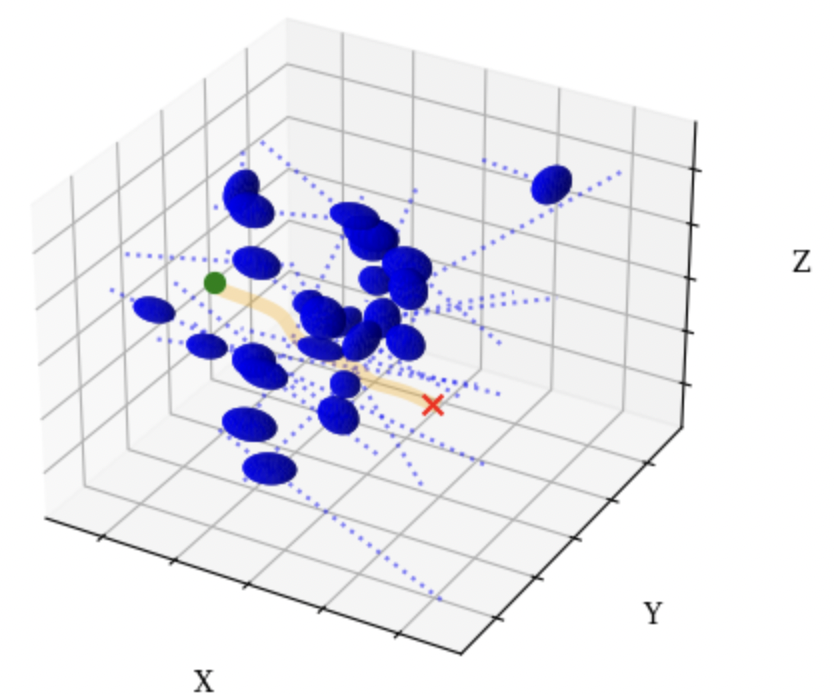}
  \caption*{b) Quadrotor}
\endminipage
\caption{Randomized comparison obstacle configurations with orange T-DBaS trajectory examples. The robot must navigate highly non-convex constraints, while the quadrotor system must navigate convex but moving obstacles with trajectories shown in blue.}
\label{fig:obstacle examples}
\vspace{-3.5mm}
\end{figure}

\begin{figure}[!htb]
\minipage{0.24\textwidth}
  \includegraphics[trim={0 0 0 0},clip,width=\linewidth]{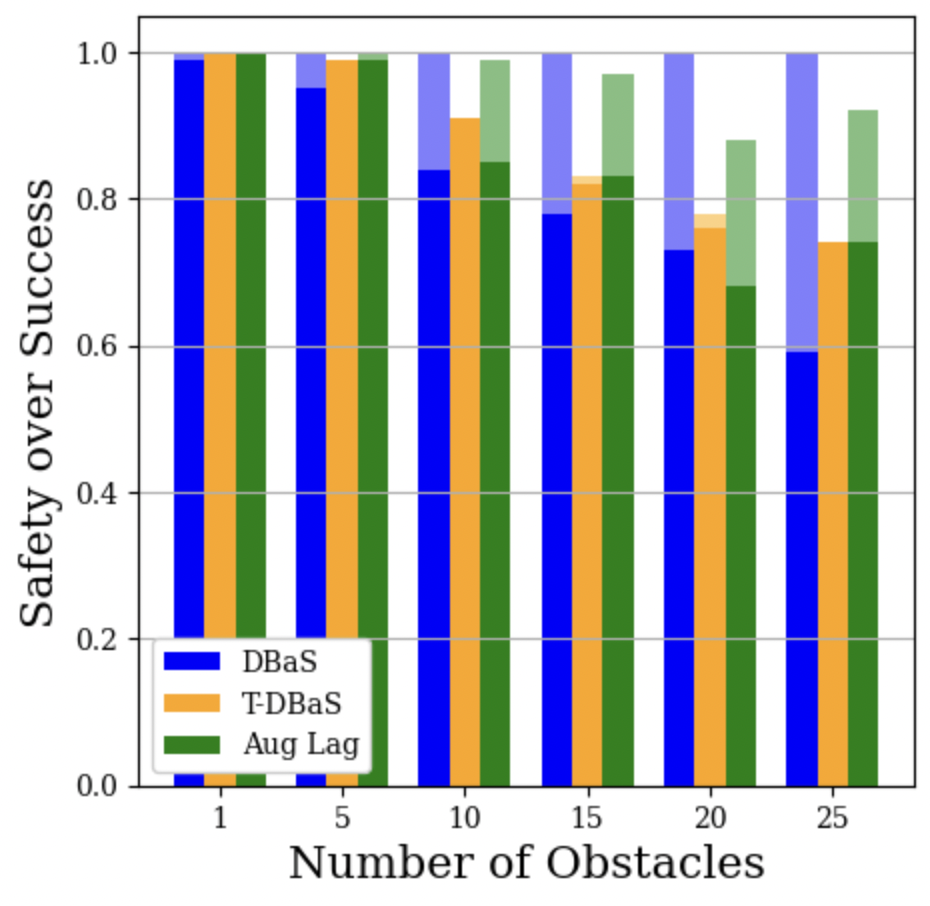}
  \caption*{a) Diff. Drive}
\endminipage\hfill
\minipage{0.24\textwidth}
  \includegraphics[trim={0 0 0 0},clip,width=\linewidth]{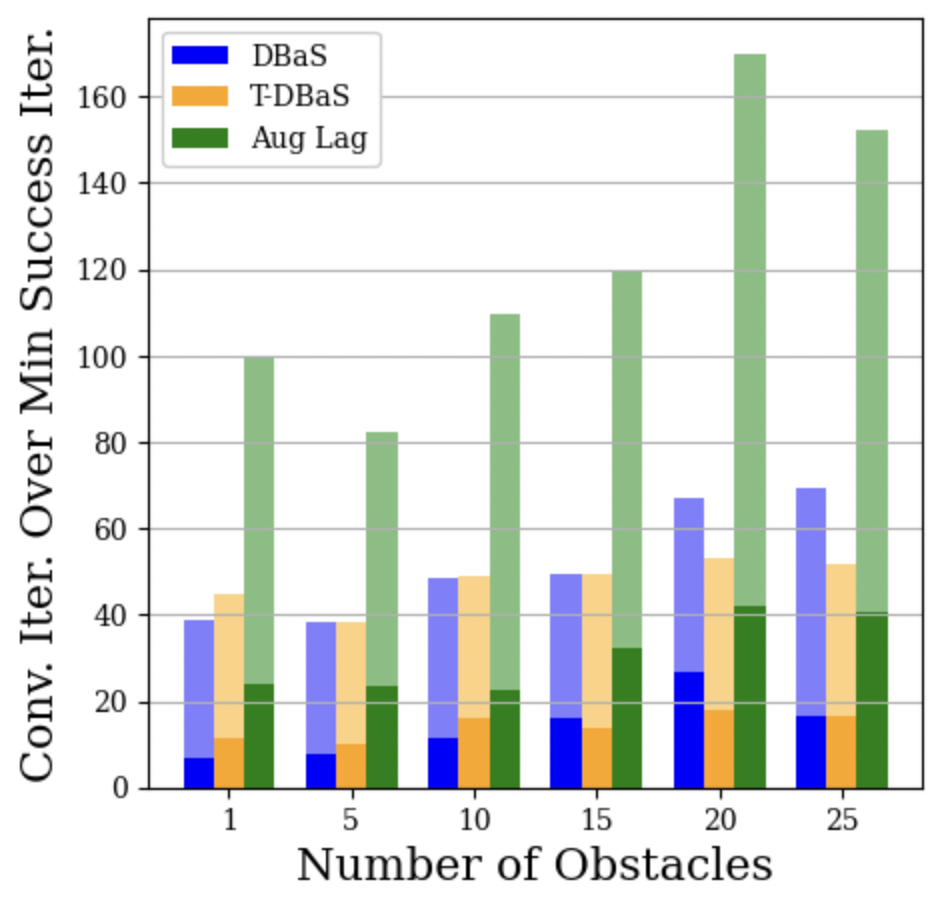}
  \caption*{b) Diff. Drive}
\endminipage\hfill
\minipage{0.24\textwidth}%
  \includegraphics[trim={0 0 0 0},clip,width=\linewidth]{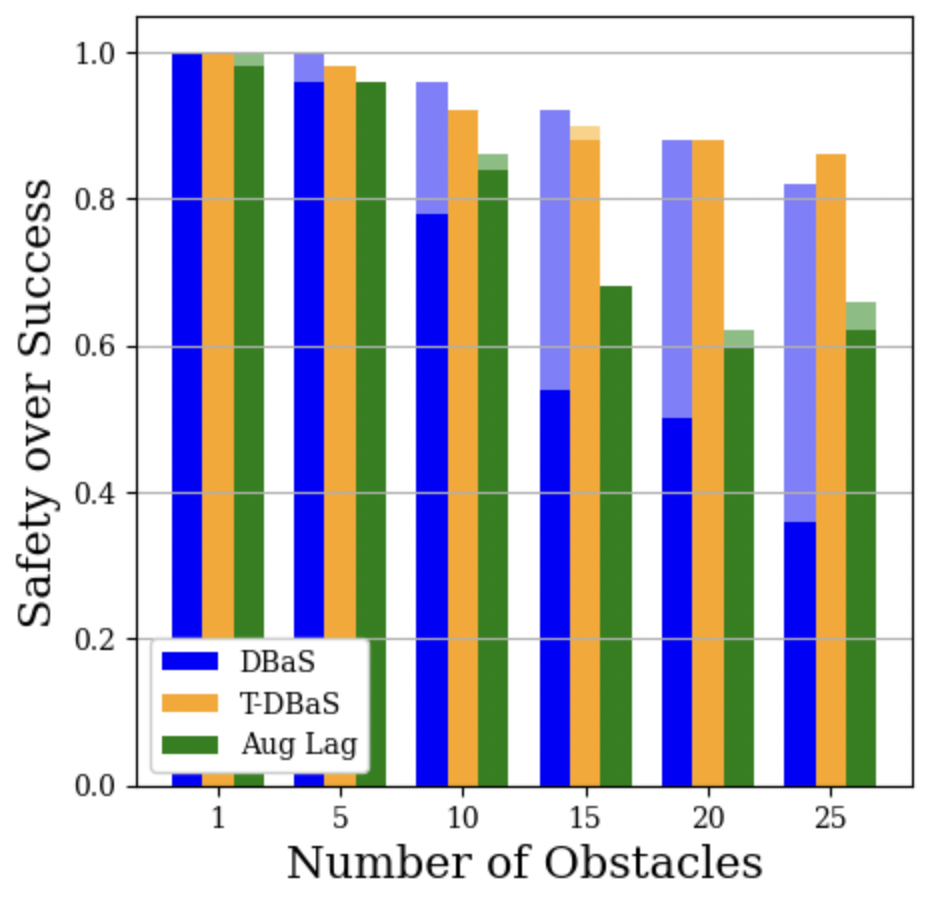}
  \caption*{c) Quadrotor}
\endminipage
\minipage{0.24\textwidth}%
  \includegraphics[trim={0 0 0 0},clip,width=\linewidth]{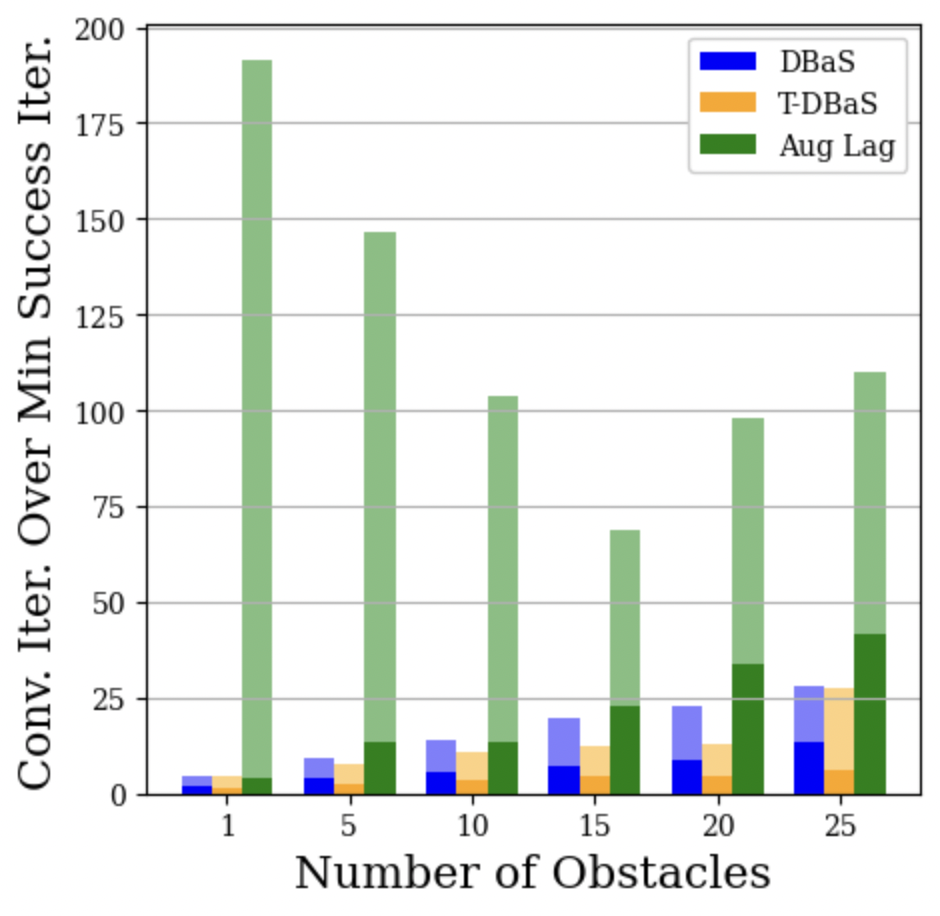}
  \caption*{d) Quadrotor}
\endminipage
\caption{Comparison results. a) and c) show average percentage of experiments that converge to safe solutions (faded) plotted above the average percentage of experiments that converge to safe and goal-reaching solutions (solid) versus number of obstacles. b) and d) show the average of minimum number iterations needed to find an initial solution that is goal-reaching and safe (faded) plotted above average iterations needed to converge (solid) versus number of obstacles.}
\label{fig:comparison results}
\vspace{-6mm}
\end{figure}
Next, we compare the T-DBaS with its classical counterpart and AL-DDP on two different systems with randomized obstacles. For tuning, we used the hyper-parameter tuning algorithm, Tree-structured Parzen Estimator (TPE) \cite{bergstra2011algorithms}, implemented in the Neural Network Intelligence (NNI) Python package \cite{nni2021}. The score function provided to TPE is the portion of trials that were goal-reaching (within $0.25$m of the desired state) and safe. We use the subset of experiments with the highest amount of randomly generated obstacles (25 obstacles in both comparisons) to train TPE, on the assumption that the hyper-parameters that perform well with many obstacles generalize to less obstacles as well.

Our implementation of AL includes zero-intialized dual variables and a schedule that reduces the iLQR convergence criteria when the resulting trajectory does not violate the constraints until that inner-loop convergence criteria is smaller than the static outer-loop convergence criteria. 

\subsubsection{Differential Drive Comparison}
For the differential drive system, we compared the three algorithms in a total of 600 experiments. These were split based on the number of obstacles, with \{1, 5, 10, 15, 20, 25\} obstacles at 100 experiments each. These obstacles are defined by randomly sized and rotated rectangles with safety function $h=|\frac{1}{r}p_{x,rot}+p_{y,rot}|+|\frac{1}{r}p_{x,rot}-p_{y,rot}|-s>0$, where $p_{x,rot}=(p_x-o_x)\cos{\theta}+(p_y-o_y)\sin{\theta}, \  p_{y,rot}=(p_x-o_x)\sin{\theta}-(p_y-o_y)\cos{\theta}$ are the relative coordinates of the agent's position $p_x, \ p_y$ from the obstacle center, $o_x,\ o_y$ rotated by $\theta$. In addition, $r$ and $s$ are scaling parameters. Obstacle parameters were uniformly sampled from $o_x,o_y\sim \emph{U}(-5,5)$, $s\sim \emph{U}(1,1.5)$, $r\sim \emph{U}(0,4)$, $\theta\sim \emph{U}(-\pi,\pi)$. The initial and goal states were also uniformly distributed within the left-hand side and the right-hand side of the obstacle field, respectively, i.e. $p_{x,0}\sim \emph{U}(-5,0)$,$p_{x,f}\sim \emph{U}(0,5)$, $p_{y,0}, \ p_{y,f} \sim \emph{U}(-5,5)$. See \autoref{fig:obstacle examples} a) for a random experiment example. The agent was given 300 time steps with $dt=0.01$s, control limits of $u_{1,2}\in(-100,100)$ enforced via clipping to avoid any abuses of discretization, a maximum of 500 DDP iterations, and a convergence criterion of when the cost reduction is less than $10^{-3}$. 

For all algorithms, the terminal cost matrix was kept consistent at $S_{1:3,1:3}=\text{diag}(500,500,0)$ with $S_{4,4}=0.05$ for the barrier state methods. For T-DBaS, the optimized parameters were $\{p,m,c_1,c_2,Q_{xy},Q_\gamma,R\}$, where $Q_{xy},Q_\gamma$ are the diagonal components for the running cost, and $R$ is the diagonal control cost. The running cost of the barrier state was held constant at $Q_{BaS}=10^{-2}$. For DBaS, $\{Q_{BaS}, Q_{xy}, Q_\gamma, R\}$ were the parameters. For AL, the optimization parameters were $\{M,\rho_0,\beta_\rho,\omega_0,\beta_\omega, Q_{xy},Q_\gamma,R\}$, where $M$ is the maximum iterations the inner loop of DDP can take, $\rho_0$ is the initial penalty, $\beta_\rho$ is the update rate that increases $\rho$ when DDP converges and the constraints are not satisfied, $\omega_0$ is the initial convergence threshold for the inner loop of DDP, and $\beta_\omega$ is the update rate that reduces $\omega$ when DDP converges with the constraints satisfied.  

The optimal parameters found by TPE for DBaS were $Q_{x,y}=1.18 \cdot 10^{-3}$, $Q_{\gamma}=2.27 \cdot 10^{-3}$, $Q_{dbas}=7.24 \cdot 10^{-4}$, and $R=9.42 \cdot 10^{-5}$. For T-DBaS they were $Q_{x,y}=1.04 \cdot 10^{-5}$, $Q_{\gamma}=4.13 \cdot 10^{-3}$, $R=1.9 \cdot 10^{-5}$, $p=21.0$, $m=10.2$, $c_1=44.8$, and $c_2=6.86$. For AL they were $Q_{x,y}=1.49 \cdot 10^{-5}$, $Q_{\gamma} = 4.12 \cdot 10^{-4}$, $M=150$, $\rho_0=33.7$, $\beta_\rho=1.18$, $\omega_0=2.77$ and $\beta_\omega=0.33$.

 \autoref{fig:comparison results} a) and b) show the results of this comparison. Firstly, by design, we see that the DBaS method always avoids collision with obstacles at convergence. T-DBaS is shown to greatly improve upon the DBaS results and rival the performance of the well-known AL method while greatly outperforming AL in both the average iterations required to reach a safe and goal-reaching solution and the average iterations required to converge. This is due in part to the static, penalty-like cost that T-DBaS enjoys, which better takes advantage of the convergence capabilities of DDP. The slow convergence of AL is potentially due to the control clipping that we implemented on all algorithms as a result of AL abusing time discretization to \textit{jump} over obstacles. Both T-DBaS and DBaS do not display this behavior. Additionally, because TPE optimized for the parameters best suited for the 25 obstacle experiments, the T-DBaS optimal parameters generalized better for lower amounts of obstacles, shown by its favorability in the 10 and 20 obstacle experiments.  Finally, we note the equal priority T-DBaS gives to safety and goal-reaching, shown by the minute difference between its two metrics in \autoref{fig:comparison results} a). 

 \begin{figure}[t]
\vspace{0mm}
    \centering
    \includegraphics[trim=2mm 0 0 0, clip,width=0.8\linewidth]{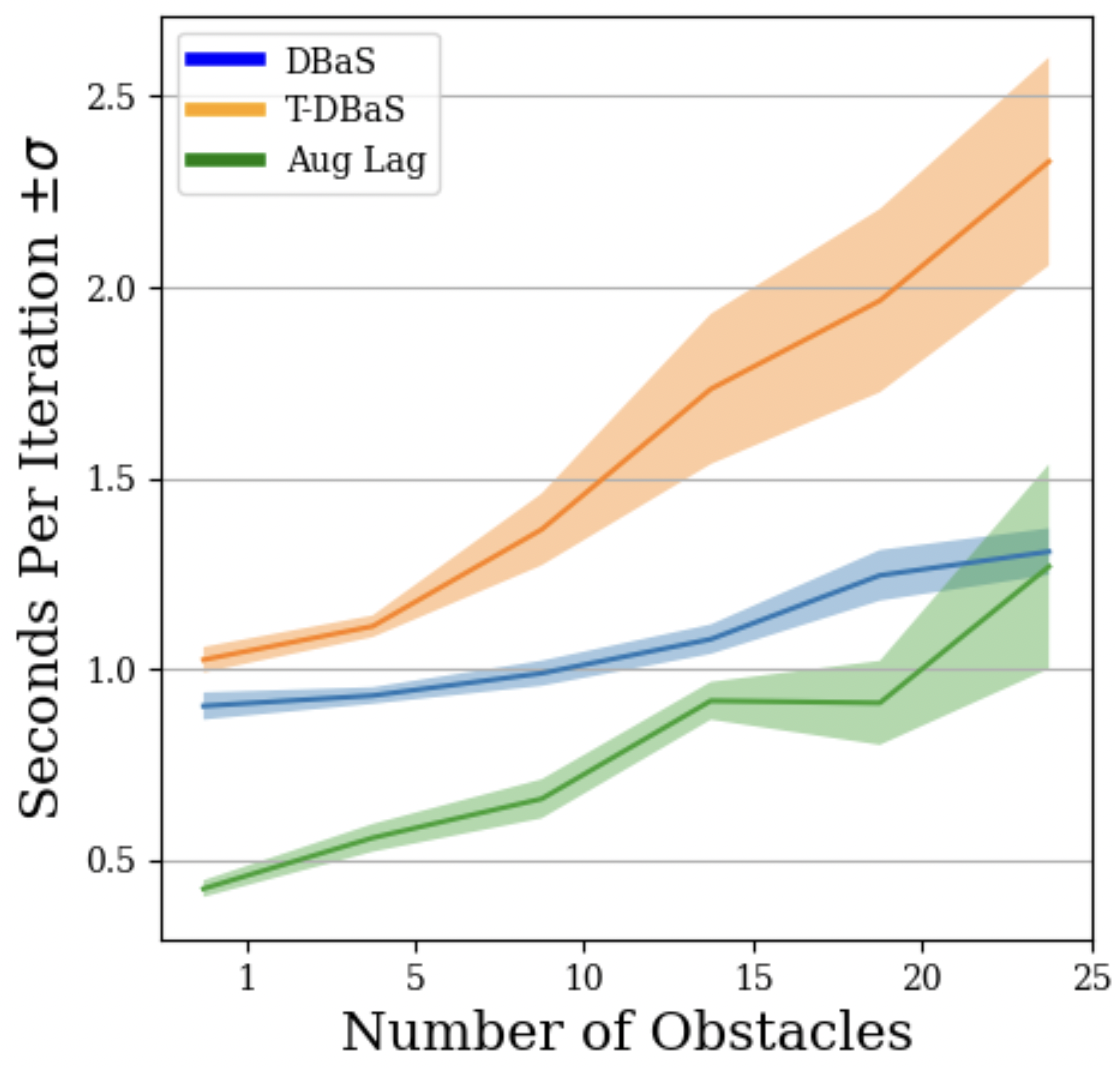} \\
    \caption{Seconds per iteration with standard deviation for each algorithm collected on 10 differential drive scenarios per data point. Collected on a M1 processor. } 
\vspace{-5mm}
\label{fig: timing}
\end{figure}

 Finally, \autoref{fig: timing} shows the time required per iteration for each algorithm. As one might expect, the more complex tolerant barrier function induces a higher computational cost by a factor of $\sim$ 2 compared to the Augmented-Lagrangian method. Note that the classical barrier function scales more efficiently with amount of constraints, likely as a result of its simplicity (as well as its derivative). Despite this, both barrier methods exhibit a convergence advantage over Augmented-Lagrangian by a factor greater than 2. 
 
 As the amount of obstacles increase, the standard deviation of the amount of time required per iteration increases only for T-DBaS and Augmented Lagrangian. This is due to these algorithms' access to the high penalties present in the unsafe region, different for each scenario, which increases computation time for matrix multiplication.
 
 From this experiment, we conclude that barrier state methods are well suited for online, model predictive control based optimization where each iteration carries more responsibility in finding the solution, as few iterations are performed per time step. Furthermore, T-DBaS excels when large amounts of non-smooth obstacles and local minima are present.

\subsubsection{Quadrotor Comparison}
For the quadrotor system, we have 50 experiments per each amount of obstacles, giving a total of 300 experiments. The obstacles are defined as ellipsoids moving periodically on a linear path between two randomly chosen points. At a single instant in time, the safety function assigned to each ellipsoid is $h=\frac{p_{x,rel}^2}{a^2}+\frac{p_{y,rel}^2}{b^2}+\frac{p_{z,rel}^2}{c^2}-1>0$, where $p_{x,rel},p_{y,rel},p_{z,rel}$ are the relative position of the agent w.r.t. the center of the obstacle. The movement of the center of the obstacles are defined with $p_{o}(t)=p_{o,0}+\vec{l}(-\cos(2\pi t/P)+1)/2$, where $p_{o,0}$, $p_{o,f}=p_{o,0}+\vec{l}$ are the two ends of the linear path and $P$ is the period of movement. The defining parameters of the obstacles are sampled uniformly within $p_{o,0},p_{o,f}\sim \emph{U}(-5,5)$ in each axis, $a,b,c\sim \emph{U}(0.3,0.7)$, $P\sim \emph{U}(5,10)$. The start and goal states were sampled within the bounds of the experiment with the start in the $-x$ region and the goal in the $+x$ region. See \autoref{fig:obstacle examples} b) for a random experiment example. The starting state was checked explicitly for constraint violations of the rollout of the nominal control initialization. The agent was given $333$ time steps with $dt=0.03$, $500$ DDP iterations, and a convergence threshold of $10^{-2}$. 

For all algorithms, the terminal cost matrix was such that $S_{1:12,1:12}=\text{diag}(0,...,0,250,250,250,500,500,500)$ with $S_{13,13}=0.05$ for the BaS methods. Also, the running cost for the yaw angle $Q_\psi$ is set as zero. For the quadrotor system, the running cost parameters to optimize were defined as $\{Q_{\phi,\theta},Q_{\dot\phi,\dot\theta,\dot\psi},Q_v,Q_p\}$, where $Q_{\phi,\theta}$ is for the pitch and roll angles, $Q_{\dot\phi,\dot\theta,\dot\psi}$ is for all angle rates, $Q_v$ is for the velocities, and $Q_p$ is for the positions. The running cost of the tolerant barrier state was held constant at $10^{-3}$. All other hyper-parameters were held the same as the previous comparisons. 

The optimal parameters found by TPE for DBaS were $Q_{\phi,\theta}=4.71$, $Q_{\dot\phi,\dot\theta,\dot\psi}=113$, $Q_v=0.0221$, $Q_p=0.0170$, $Q_{dbas}=0.0034$, and $R=40.64$. For T-DBaS they were $Q_{\phi,\theta}=1.26 \cdot 10^{-5}$, $Q_{\dot\phi,\dot\theta,\dot\psi}=5.57 \cdot 10^{-4}$, $Q_v=0.299$, $Q_p=5.90 \cdot 10^{-3}$, $R=0.509$, $p=117$, $m=97.9$, $c_1=11.6$, and $c_2=32.9$. For AL, they were $Q_{\phi,\theta}=0.0234$, $Q_{\dot\phi,\dot\theta,\dot\psi}=256$, $Q_v=34.5$, $Q_p=21.5$, $R=20.1$, $M=150$, $\rho_0=5.48$, $\beta_\rho=2.31$, $\omega_0=4.42$, and $\beta_\omega=0.305$.

Due to the inherent instability of the quadrotor system, we found that many points in the hyper-parameter search space resulted in divergence due to a loss of stability. As a result, we found it necessary to exit the barrier state method trials as soon as a single experiment resulted in divergence, as this helped the TPE algorithm find a section of the search space that resulted in high rates of goal-reaching and safety. For the AL algoirithm, we chose to penalize divergent behavior in the score function provided to TPE, in addition to exiting any trials early that resulted in 5 divergent experiments. We found this more lenient penalization and experiment cancelation schedule necessary, as TPE initially found it difficult to search for hyper-parameters that resulted in a trial with <5 divergent experiments. This is in contrast to how the stability properties of the barrier state methods were largely robust to the hyper-parameter choice. This likely affected the parts of the AL hyper-parameter search-space that TPE was biased towards, as there are areas with a higher likelihood of divergence, but also a higher rate of goal-reaching and safety. 

Subfigures c) and d) of \autoref{fig:comparison results} show the results of this comparison. Here, we see the DBaS method struggle to find safe solutions as a result of line search failure in local minima. However, its exploratory counterpart does not suffer from these minima. We hypothesize that the out-performance of T-DBaS over AL in this case is a result of the dynamical stability of T-DBaS-DDP over AL-DDP, especially given the highly mobile obstacles present within the experiments. 

  


\section{Conclusions and Future Work}  \label{CONCLUSIONS}

In this paper, we proposed T-DBaS, a variation of DBaS with increased exploration capabilities while approximating the safety guarantees of DBaS. The T-DBaS was designed such that its value increases inside the unsafe regions and retains its derivative information which is then leveraged to find non-trivial solutions. Subsequently, we embedded T-DBaS into the DDP framework for optimal control, and developed T-DBaS-DDP for safe trajectory optimization. The proposed algorithm T-DBaS-DDP was then implemented on a differential drive robot and a quadrotor in various examples with experimentation on a multi-robot hardware platform. We compared the proposed algorithm against its classical counterpart, DBaS-DDP, as well as against AL-DDP. An interesting future direction is learning the tolerant barrier function parameters according to state and obstacle uncertainty via differentiable optimal control. 
\vspace{-2mm}






\printbibliography

@article{Peng2009barrier-lyap,
title = {Barrier Lyapunov Functions for the control of output-constrained nonlinear systems},
journal = {Automatica},
volume = {45},
number = {4},
pages = {918-927},
year = {2009},
author = {Keng Peng Tee and Shuzhi Sam Ge and Eng Hock Tay},
}

@article{mayne1966second,
  title={A second-order gradient method for determining optimal trajectories of non-linear discrete-time systems},
  author={Mayne, David},
  journal={International Journal of Control},
  volume={3},
  number={1},
  pages={85--95},
  year={1966},
  publisher={Taylor \& Francis}
}

@book{jacobson1970differential,
  title={Differential dynamic programming},
  author={Jacobson, David H and Mayne, David Q},
  year={1970},
  publisher={North-Holland}
}

@article{aoyama2020constrainedDDP,
  title={\href{https://arxiv.org/abs/2005.00985}{Constrained Differential Dynamic Programming Revisited}},
  author={Aoyama, Yuichiro and Boutselis, George and Patel, Akash and Theodorou, Evangelos A},
  journal={arXiv preprint arXiv:2005.00985},
  year={2020}
}

@inproceedings{tassa2012synthesis,
  title={Synthesis and stabilization of complex behaviors through online trajectory optimization},
  author={Tassa, Yuval and Erez, Tom and Todorov, Emanuel},
  booktitle={2012 IEEE/RSJ International Conference on Intelligent Robots and Systems},
  pages={4906--4913},
  year={2012},
  organization={IEEE}
}

@ARTICLE{Almubarak2021SafetyEC,
  author={Almubarak, Hassan and Sadegh, Nader and Theodorou, Evangelos A.},
  journal={IEEE Control Systems Letters}, 
  title={{Safety Embedded Control of Nonlinear Systems via Barrier States}}, 
  year={2022},
  volume={6},
  number={},
  pages={1328--1333}}

@inproceedings{prajna2004safety,
  title={{Safety verification of hybrid systems using barrier certificates}},
  author={Prajna, Stephen and Jadbabaie, Ali},
  booktitle={International Workshop on Hybrid Systems: Computation and Control},
  pages={477--492},
  year={2004},
  organization={Springer}
}

@article{ames2016CBF-forSaferyCritControl,
  title={{Control barrier function based quadratic programs for safety critical systems}},
  author={Ames, Aaron D and Xu, Xiangru and Grizzle, Jessy W and Tabuada, Paulo},
  journal={IEEE Transactions on Automatic Control},
  volume={62},
  number={8},
  pages={3861--3876},
  year={2016},
  publisher={IEEE}
}

@article{romdlony2016stabilization,
  title={Stabilization with guaranteed safety using control Lyapunov--barrier function},
  author={Romdlony, Muhammad Zakiyullah and Jayawardhana, Bayu},
  journal={Automatica},
  volume={66},
  pages={39--47},
  year={2016},
  publisher={Elsevier}
}

@article{wieland2007constructive,
  title={{Constructive safety using control barrier functions}},
  author={Wieland, Peter and Allg{\"o}wer, Frank},
  journal={IFAC Proceedings Volumes},
  volume={40},
  number={12},
  pages={462--467},
  year={2007},
  publisher={Elsevier}
}

@inproceedings{Sabatino2015QuadrotorCM,
  title={{Quadrotor control: modeling, nonlinearcontrol design, and simulation}},
  author={F. Sabatino},
  year={2015}
}

@INPROCEEDINGS{howell2019altro,
author={T. A. {Howell} and B. E. {Jackson} and Z. {Manchester}},
booktitle={2019 IEEE/RSJ International Conference on Intelligent Robots and Systems (IROS)},   
title={\href{https://ieeexplore.ieee.org/document/8967788}{ALTRO: A Fast Solver for Constrained Trajectory Optimization}},
year={2019},
volume={},
number={},
pages={7674-7679}
}

@ARTICLE{almubarak2022safeddp,
  author={Almubarak, Hassan and Stachowicz, Kyle and Sadegh, Nader and Theodorou, Evangelos A.},
  journal={IEEE Robotics and Automation Letters}, 
  title={{Safety Embedded Differential Dynamic Programming using Discrete Barrier States}},
  year={2022},
  volume={7},
  number={2},
  pages={2755-2762}}

@Inbook{diehl2009nonlinearmpc,
author="Diehl, Moritz
and Ferreau, Hans Joachim
and Haverbeke, Niels",
title="Efficient Numerical Methods for Nonlinear MPC and Moving Horizon Estimation",
bookTitle="Nonlinear Model Predictive Control: Towards New Challenging Applications",
year="2009",
publisher="Springer Berlin Heidelberg",
pages="391--417",
isbn="978-3-642-01094-1"
}

@article{wilson2020robotarium,
  title={The robotarium: Globally impactful opportunities, challenges, and lessons learned in remote-access, distributed control of multirobot systems},
  author={Wilson, Sean and Glotfelter, Paul and Wang, Li and Mayya, Siddharth and Notomista, Gennaro and Mote, Mark and Egerstedt, Magnus},
  journal={IEEE Control Systems Magazine},
  volume={40},
  number={1},
  pages={26--44},
  year={2020},
  publisher={IEEE}
}

@article{von1992direct,
  title={Direct and indirect methods for trajectory optimization},
  author={Von Stryk, Oskar and Bulirsch, Roland},
  journal={Annals of operations research},
  volume={37},
  number={1},
  pages={357--373},
  year={1992},
  publisher={Springer}
}

@INPROCEEDINGS{kumar2016optimal,
  author={Kumar, Vikash and Todorov, Emanuel and Levine, Sergey},
  booktitle={2016 IEEE International Conference on Robotics and Automation (ICRA)}, 
  title={Optimal control with learned local models: Application to dexterous manipulation}, 
  year={2016},
  volume={},
  number={},
  pages={378-383},
}

@article{jallet2022constrained,
  title={Constrained Differential Dynamic Programming: A primal-dual augmented Lagrangian approach},
  author={Jallet, Wilson and Bambade, Antoine and Mansard, Nicolas and Carpentier, Justin},
  year={2022}
}

@inproceedings{aoyama2021receding,
  title={Receding Horizon Differential Dynamic Programming Under Parametric Uncertainty},
  author={Aoyama, Yuichiro and Saravanos, Augustinos D and Theodorou, Evangelos A},
  booktitle={2021 60th IEEE Conference on Decision and Control (CDC)},
  pages={3761--3767},
  year={2021},
  organization={IEEE}
}

@article{bergstra2011algorithms,
  title={Algorithms for hyper-parameter optimization},
  author={Bergstra, James and Bardenet, R{\'e}mi and Bengio, Yoshua and K{\'e}gl, Bal{\'a}zs},
  journal={Advances in neural information processing systems},
  volume={24},
  year={2011}
}

@software{nni2021,
   author = {{Microsoft}},
   month = {1},
   title = {{Neural Network Intelligence}},
   url = {https://github.com/microsoft/nni},
   version = {2.0},
   year = {2021}
}

@inproceedings{alothman2016quadrotor,
  title={Quadrotor transporting cable-suspended load using iterative linear quadratic regulator (ilqr) optimal control},
  author={Alothman, Yaser and Gu, Dongbing},
  booktitle={2016 8th Computer Science and Electronic Engineering (CEEC)},
  pages={168--173},
  year={2016},
  organization={IEEE}
}

@ARTICLE{saravanos2022distributed,
  author={Saravanos, Augustinos D. and Aoyama, Yuichiro and Zhu, Hongchang and Theodorou, Evangelos A.},
  journal={IEEE Transactions on Robotics}, 
  title={Distributed Differential Dynamic Programming Architectures for Large-Scale Multiagent Control}, 
  year={2023},
  volume={},
  number={},
  pages={1-21},
  doi={10.1109/TRO.2023.3319894}}

\end{document}